\documentclass{edm_template} 
\usepackage{amsmath,epsfig} 
\usepackage{amsfonts}
\usepackage{amssymb}
\usepackage{mathrsfs}
\usepackage{subfigure}
\usepackage{paralist}
\usepackage{booktabs}
\usepackage{color}
\DeclareGraphicsExtensions{.pdf,.jpeg,.png,.epbibdesks}
\usepackage[T1]{fontenc}
\usepackage{epstopdf}
\usepackage{graphicx}

\usepackage{floatrow}
\newfloatcommand{capbtabbox}{table}[][\FBwidth]
\usepackage[algo2e]{algorithm2e}
\usepackage{yhmath}
\usepackage{amssymb}
\usepackage{amsfonts}
\usepackage{mathrsfs}
\usepackage{xspace}
\usepackage{bm}
\usepackage{upgreek}

\newcommand{\safemath}[2]{\newcommand{#1}{\ensuremath{#2}\xspace}}

\safemath{\bma}{\mathbf{a}}
\safemath{\bmb}{\mathbf{b}}
\safemath{\bmc}{\mathbf{c}}
\safemath{\bmd}{\mathbf{d}}
\safemath{\bme}{\mathbf{e}}
\safemath{\bmf}{\mathbf{f}}
\safemath{\bmg}{\mathbf{g}}
\safemath{\bmh}{\mathbf{h}}
\safemath{\bmi}{\mathbf{i}}
\safemath{\bmj}{\mathbf{j}}
\safemath{\bmk}{\mathbf{k}}
\safemath{\bml}{\mathbf{l}}
\safemath{\bmm}{\mathbf{m}}
\safemath{\bmn}{\mathbf{n}}
\safemath{\bmo}{\mathbf{o}}
\safemath{\bmp}{\mathbf{p}}
\safemath{\bmq}{\mathbf{q}}
\safemath{\bmr}{\mathbf{r}}
\safemath{\bms}{\mathbf{s}}
\safemath{\bmt}{\mathbf{t}}
\safemath{\bmu}{\mathbf{u}}
\safemath{\bmv}{\mathbf{v}}
\safemath{\bmw}{\mathbf{w}}
\safemath{\bmx}{\mathbf{x}}
\safemath{\bmy}{\mathbf{y}}
\safemath{\bmz}{\mathbf{z}}
\safemath{\bmzero}{\mathbf{0}}
\safemath{\bmone}{\mathbf{1}}

\bmdefine{\biad}{a}
\bmdefine{\bibd}{b}
\bmdefine{\bicd}{c}
\bmdefine{\bidd}{d}
\bmdefine{\bied}{e}
\bmdefine{\bifd}{f}
\bmdefine{\bigd}{g}
\bmdefine{\bihd}{h}
\bmdefine{\biid}{i}
\bmdefine{\bijd}{j}
\bmdefine{\bikd}{k}
\bmdefine{\bild}{l}
\bmdefine{\bimd}{m}
\bmdefine{\bind}{n}
\bmdefine{\biod}{o}
\bmdefine{\bipd}{p}
\bmdefine{\biqd}{q}
\bmdefine{\bird}{r}
\bmdefine{\bisd}{s}
\bmdefine{\bitd}{t}
\bmdefine{\biud}{u}
\bmdefine{\bivd}{v}
\bmdefine{\biwd}{w}
\bmdefine{\bixd}{x}
\bmdefine{\biyd}{y}
\bmdefine{\bizd}{z}

\bmdefine{\bixid}{\xi}
\bmdefine{\bilambdad}{\lambda}
\bmdefine{\bimud}{\mu}
\bmdefine{\bithetad}{\theta}
\bmdefine{\biphid}{\phi}
\bmdefine{\bideltad}{\delta}

\safemath{\bmia}{\biad}
\safemath{\bmib}{\bibd}
\safemath{\bmic}{\bicd}
\safemath{\bmid}{\bidd}
\safemath{\bmie}{\bied}
\safemath{\bmif}{\bifd}
\safemath{\bmig}{\bigd}
\safemath{\bmih}{\bihd}
\safemath{\bmii}{\biid}
\safemath{\bmij}{\bijd}
\safemath{\bmik}{\bikd}
\safemath{\bmil}{\bild}
\safemath{\bmim}{\bimd}
\safemath{\bmin}{\bind}
\safemath{\bmio}{\biod}
\safemath{\bmip}{\bipd}
\safemath{\bmiq}{\biqd}
\safemath{\bmir}{\bird}
\safemath{\bmis}{\bisd}
\safemath{\bmit}{\bitd}
\safemath{\bmiu}{\biud}
\safemath{\bmiv}{\bivd}
\safemath{\bmiw}{\biwd}
\safemath{\bmix}{\bixd}
\safemath{\bmiy}{\biyd}
\safemath{\bmiz}{\bizd}

\safemath{\bmxi}{\bixid}
\safemath{\bmlambda}{\bilambdad}
\safemath{\bmmu}{\bimud}
\safemath{\bmtheta}{\bithetad}
\safemath{\bmphi}{\biphid}
\safemath{\bmdelta}{\bideltad}

\safemath{\bA}{\mathbf{A}}
\safemath{\bB}{\mathbf{B}}
\safemath{\bC}{\mathbf{C}}
\safemath{\bD}{\mathbf{D}}
\safemath{\bE}{\mathbf{E}}
\safemath{\bF}{\mathbf{F}}
\safemath{\bG}{\mathbf{G}}
\safemath{\bH}{\mathbf{H}}
\safemath{\bI}{\mathbf{I}}
\safemath{\bJ}{\mathbf{J}}
\safemath{\bK}{\mathbf{K}}
\safemath{\bL}{\mathbf{L}}
\safemath{\bM}{\mathbf{M}}
\safemath{\bN}{\mathbf{N}}
\safemath{\bO}{\mathbf{O}}
\safemath{\bP}{\mathbf{P}}
\safemath{\bQ}{\mathbf{Q}}
\safemath{\bR}{\mathbf{R}}
\safemath{\bS}{\mathbf{S}}
\safemath{\bT}{\mathbf{T}}
\safemath{\bU}{\mathbf{U}}
\safemath{\bV}{\mathbf{V}}
\safemath{\bW}{\mathbf{W}}
\safemath{\bX}{\mathbf{X}}
\safemath{\bY}{\mathbf{Y}}
\safemath{\bZ}{\mathbf{Z}}

\safemath{\bZero}{\mathbf{0}}
\safemath{\bOne}{\mathbf{1}}
\safemath{\bDelta}{\mathbf{\Delta}}
\safemath{\bLambda}{\mathbf{\UpLambda}}
\safemath{\bPhi}{\mathbf{\Upphi}}
\safemath{\bSigma}{\mathbf{\Upsigma}}
\safemath{\bOmega}{\mathbf{\Upomega}}
\safemath{\bTheta}{\mathbf{\Uptheta}}

\bmdefine{\biAd}{A}
\bmdefine{\biBd}{B}
\bmdefine{\biCd}{C}
\bmdefine{\biDd}{D}
\bmdefine{\biEd}{E}
\bmdefine{\biFd}{F}
\bmdefine{\biGd}{G}
\bmdefine{\biHd}{H}
\bmdefine{\biId}{I}
\bmdefine{\biJd}{J}
\bmdefine{\biKd}{K}
\bmdefine{\biLd}{L}
\bmdefine{\biMd}{M}
\bmdefine{\biNd}{N}
\bmdefine{\biOd}{O}
\bmdefine{\biPd}{P}
\bmdefine{\biQd}{Q}
\bmdefine{\biRd}{R}
\bmdefine{\biSd}{S}
\bmdefine{\biTd}{T}
\bmdefine{\biUd}{U}
\bmdefine{\biVd}{V}
\bmdefine{\biWd}{W}
\bmdefine{\biXd}{X}
\bmdefine{\biYd}{Y}
\bmdefine{\biZd}{Z}

\bmdefine{\biDelta}{\Delta}
\bmdefine{\biLambda}{\Lambda}
\bmdefine{\biPhi}{\Phi}
\bmdefine{\biSigma}{\Sigma}
\bmdefine{\biOmega}{\Omega}
\bmdefine{\biTheta}{\Theta}

\safemath{\bimA}{\biAd}
\safemath{\bimB}{\biBd}
\safemath{\bimC}{\biCd}
\safemath{\bimD}{\biDd}
\safemath{\bimE}{\biEd}
\safemath{\bimF}{\biFd}
\safemath{\bimG}{\biGd}
\safemath{\bimH}{\biHd}
\safemath{\bimI}{\biId}
\safemath{\bimJ}{\biJd}
\safemath{\bimK}{\biKd}
\safemath{\bimL}{\biLd}
\safemath{\bimM}{\biMd}
\safemath{\bimN}{\biNd}
\safemath{\bimO}{\biOd}
\safemath{\bimP}{\biPd}
\safemath{\bimQ}{\biQd}
\safemath{\bimR}{\biRd}
\safemath{\bimS}{\biSd}
\safemath{\bimT}{\biTd}
\safemath{\bimU}{\biUd}
\safemath{\bimV}{\biVd}
\safemath{\bimW}{\biWd}
\safemath{\bimX}{\biXd}
\safemath{\bimY}{\biYd}
\safemath{\bimZ}{\biZd}

\safemath{\bimDelta}{\biDelta}
\safemath{\bimLambda}{\biLambda}
\safemath{\bimPhi}{\biPhi}
\safemath{\bimSigma}{\biSigma}
\safemath{\bimOmega}{\biOmega}
\safemath{\bimTheta}{\biTheta}

\safemath{\setA}{\mathcal{A}}
\safemath{\setB}{\mathcal{B}}
\safemath{\setC}{\mathcal{C}}
\safemath{\setD}{\mathcal{D}}
\safemath{\setE}{\mathcal{E}}
\safemath{\setF}{\mathcal{F}}
\safemath{\setG}{\mathcal{G}}
\safemath{\setH}{\mathcal{H}}
\safemath{\setI}{\mathcal{I}}
\safemath{\setJ}{\mathcal{J}}
\safemath{\setK}{\mathcal{K}}
\safemath{\setL}{\mathcal{L}}
\safemath{\setM}{\mathcal{M}}
\safemath{\setN}{\mathcal{N}}
\safemath{\setO}{\mathcal{O}}
\safemath{\setP}{\mathcal{P}}
\safemath{\setQ}{\mathcal{Q}}
\safemath{\setR}{\mathcal{R}}
\safemath{\setS}{\mathcal{S}}
\safemath{\setT}{\mathcal{T}}
\safemath{\setU}{\mathcal{U}}
\safemath{\setV}{\mathcal{V}}
\safemath{\setW}{\mathcal{W}}
\safemath{\setX}{\mathcal{X}}
\safemath{\setY}{\mathcal{Y}}
\safemath{\setZ}{\mathcal{Z}}
\safemath{\emptySet}{\varnothing}

\safemath{\colA}{\mathscr{A}}
\safemath{\colB}{\mathscr{B}}
\safemath{\colC}{\mathscr{C}}
\safemath{\colD}{\mathscr{D}}
\safemath{\colE}{\mathscr{E}}
\safemath{\colF}{\mathscr{F}}
\safemath{\colG}{\mathscr{G}}
\safemath{\colH}{\mathscr{H}}
\safemath{\colI}{\mathscr{I}}
\safemath{\colJ}{\mathscr{J}}
\safemath{\colK}{\mathscr{K}}
\safemath{\colL}{\mathscr{L}}
\safemath{\colM}{\mathscr{M}}
\safemath{\colN}{\mathscr{N}}
\safemath{\colO}{\mathscr{O}}
\safemath{\colP}{\mathscr{P}}
\safemath{\colQ}{\mathscr{Q}}
\safemath{\colR}{\mathscr{R}}
\safemath{\colS}{\mathscr{S}}
\safemath{\colT}{\mathscr{T}}
\safemath{\colU}{\mathscr{U}}
\safemath{\colV}{\mathscr{V}}
\safemath{\colW}{\mathscr{W}}
\safemath{\colX}{\mathscr{X}}
\safemath{\colY}{\mathscr{Y}}
\safemath{\colZ}{\mathscr{Z}}

\safemath{\opA}{\mathbb{A}}
\safemath{\opB}{\mathbb{B}}
\safemath{\opC}{\mathbb{C}}
\safemath{\opD}{\mathbb{D}}
\safemath{\opE}{\mathbb{E}}
\safemath{\opF}{\mathbb{F}}
\safemath{\opG}{\mathbb{G}}
\safemath{\opH}{\mathbb{H}}
\safemath{\opI}{\mathbb{I}}
\safemath{\opJ}{\mathbb{J}}
\safemath{\opK}{\mathbb{K}}
\safemath{\opL}{\mathbb{L}}
\safemath{\opM}{\mathbb{M}}
\safemath{\opN}{\mathbb{N}}
\safemath{\opO}{\mathbb{O}}
\safemath{\opP}{\mathbb{P}}
\safemath{\opQ}{\mathbb{Q}}
\safemath{\opR}{\mathbb{R}}
\safemath{\opS}{\mathbb{S}}
\safemath{\opT}{\mathbb{T}}
\safemath{\opU}{\mathbb{U}}
\safemath{\opV}{\mathbb{V}}
\safemath{\opW}{\mathbb{W}}
\safemath{\opX}{\mathbb{X}}
\safemath{\opY}{\mathbb{Y}}
\safemath{\opZ}{\mathbb{Z}}
\safemath{\opZero}{\mathbb{O}}
\safemath{\identityop}{\opI}

\safemath{\veca}{\bma}
\safemath{\vecb}{\bmb}
\safemath{\vecc}{\bmc}
\safemath{\vecd}{\bmd}
\safemath{\vece}{\bme}
\safemath{\vecf}{\bmf}
\safemath{\vecg}{\bmg}
\safemath{\vech}{\bmh}
\safemath{\veci}{\bmi}
\safemath{\vecj}{\bmj}
\safemath{\veck}{\bmk}
\safemath{\vecl}{\bml}
\safemath{\vecm}{\bmm}
\safemath{\vecn}{\bmn}
\safemath{\veco}{\bmo}
\safemath{\vecp}{\bmp}
\safemath{\vecq}{\bmq}
\safemath{\vecr}{\bmr}
\safemath{\vecs}{\bms}
\safemath{\vect}{\bmt}
\safemath{\vecu}{\bmu}
\safemath{\vecv}{\bmv}
\safemath{\vecw}{\bmw}
\safemath{\vecx}{\bmx}
\safemath{\vecy}{\bmy}
\safemath{\vecz}{\bmz}

\safemath{\veczero}{\bmzero}
\safemath{\vecone}{\bmone}
\safemath{\vecxi}{\bmxi}
\safemath{\veclambda}{\bmlambda}
\safemath{\vecmu}{\bmmu}
\safemath{\vectheta}{\bmtheta}
\safemath{\vecphi}{\bmphi}
\safemath{\vecdelta}{\bmdelta}

\safemath{\matA}{\bA}
\safemath{\matB}{\bB}
\safemath{\matC}{\bC}
\safemath{\matD}{\bD}
\safemath{\matE}{\bE}
\safemath{\matF}{\bF}
\safemath{\matG}{\bG}
\safemath{\matH}{\bH}
\safemath{\matI}{\bI}
\safemath{\matJ}{\bJ}
\safemath{\matK}{\bK}
\safemath{\matL}{\bL}
\safemath{\matM}{\bM}
\safemath{\matN}{\bN}
\safemath{\matO}{\bO}
\safemath{\matP}{\bP}
\safemath{\matQ}{\bQ}
\safemath{\matR}{\bR}
\safemath{\matS}{\bS}
\safemath{\matT}{\bT}
\safemath{\matU}{\bU}
\safemath{\matV}{\bV}
\safemath{\matW}{\bW}
\safemath{\matX}{\bX}
\safemath{\matY}{\bY}
\safemath{\matZ}{\bZ}
\safemath{\matzero}{\bmzero}

\safemath{\matDelta}{\bDelta}
\safemath{\matLambda}{\bLambda}
\safemath{\matPhi}{\bPhi}
\safemath{\matSigma}{\bSigma}
\safemath{\matOmega}{\bOmega}
\safemath{\matTheta}{\bTheta}

\safemath{\matidentity}{\matI}
\safemath{\matone}{\matO}

\safemath{\rnda}{A}
\safemath{\rndb}{B}
\safemath{\rndc}{C}
\safemath{\rndd}{D}
\safemath{\rnde}{E}
\safemath{\rndf}{F}
\safemath{\rndg}{G}
\safemath{\rndh}{H}
\safemath{\rndi}{I}
\safemath{\rndj}{J}
\safemath{\rndk}{K}
\safemath{\rndl}{L}
\safemath{\rndm}{M}
\safemath{\rndn}{N}
\safemath{\rndo}{O}
\safemath{\rndp}{P}
\safemath{\rndq}{Q}
\safemath{\rndr}{R}
\safemath{\rnds}{S}
\safemath{\rndt}{T}
\safemath{\rndu}{U}
\safemath{\rndv}{V}
\safemath{\rndw}{W}
\safemath{\rndx}{X}
\safemath{\rndy}{Y}
\safemath{\rndz}{Z}

\safemath{\rveca}{\bimA}
\safemath{\rvecb}{\bimB}
\safemath{\rvecc}{\bimC}
\safemath{\rvecd}{\bimD}
\safemath{\rvece}{\bimE}
\safemath{\rvecf}{\bimF}
\safemath{\rvecg}{\bimG}
\safemath{\rvech}{\bimH}
\safemath{\rveci}{\bimI}
\safemath{\rvecj}{\bimJ}
\safemath{\rveck}{\bimK}
\safemath{\rvecl}{\bimL}
\safemath{\rvecm}{\bimM}
\safemath{\rvecn}{\bimN}
\safemath{\rveco}{\bomO}
\safemath{\rvecp}{\bimP}
\safemath{\rvecq}{\bimQ}
\safemath{\rvecr}{\bimR}
\safemath{\rvecs}{\bimS}
\safemath{\rvect}{\bimT}
\safemath{\rvecu}{\bimU}
\safemath{\rvecv}{\bimV}
\safemath{\rvecw}{\bimW}
\safemath{\rvecx}{\bimX}
\safemath{\rvecy}{\bimY}
\safemath{\rvecz}{\bimZ}

\safemath{\rvecxi}{\bmxi}
\safemath{\rveclambda}{\bmlambda}
\safemath{\rvecmu}{\bmmu}
\safemath{\rvectheta}{\bmtheta}
\safemath{\rvecphi}{\bmphi}

\safemath{\rmatA}{\bimA}
\safemath{\rmatB}{\bimB}
\safemath{\rmatC}{\bimC}
\safemath{\rmatD}{\bimD}
\safemath{\rmatE}{\bimE}
\safemath{\rmatF}{\bimF}
\safemath{\rmatG}{\bimG}
\safemath{\rmatH}{\bimH}
\safemath{\rmatI}{\bimI}
\safemath{\rmatJ}{\bimJ}
\safemath{\rmatK}{\bimK}
\safemath{\rmatL}{\bimL}
\safemath{\rmatM}{\bimM}
\safemath{\rmatN}{\bimN}
\safemath{\rmatO}{\bimO}
\safemath{\rmatP}{\bimP}
\safemath{\rmatQ}{\bimQ}
\safemath{\rmatR}{\bimR}
\safemath{\rmatS}{\bimS}
\safemath{\rmatT}{\bimT}
\safemath{\rmatU}{\bimU}
\safemath{\rmatV}{\bimV}
\safemath{\rmatW}{\bimW}
\safemath{\rmatX}{\bimX}
\safemath{\rmatY}{\bimY}
\safemath{\rmatZ}{\bimZ}

\safemath{\rmatDelta}{\bimDelta}
\safemath{\rmatLambda}{\bimLambda}
\safemath{\rmatPhi}{\bimPhi}
\safemath{\rmatSigma}{\bimSigma}
\safemath{\rmatOmega}{\bimOmega}
\safemath{\rmatTheta}{\bimTheta}

\usepackage{amssymb}
\usepackage{amsfonts}
\usepackage{mathrsfs}
\usepackage{xspace}
\usepackage{bm}
\usepackage{fancyref}
\usepackage{textcomp}

\usepackage{multirow}
\usepackage{stmaryrd}

\newenvironment{textbmatrix}{	\setlength{\arraycolsep}{2.5pt}%
								\big[\begin{matrix}}{\end{matrix}\big]%
								\raisebox{0.08ex}{\vphantom{M}}}

\def\be{\begin{equation}}
\def\ee{\end{equation}}
\def\een{\nonumber \end{equation}}
\def\mat{\begin{bmatrix}}
\def\emat{\end{bmatrix}}
\def\btm{\begin{textbmatrix}}
\def\etm{\end{textbmatrix}}

\def\ba#1\ea{\begin{align}#1\end{align}}
\def\bas#1\eas{\begin{align*}#1\end{align*}}
\def\bs#1\es{\begin{split}#1\end{split}} 
\def\bg#1\eg{\begin{gather}#1\end{gather}}
\def\bml#1\eml{\begin{multline}#1\end{multline}}
\def\bi#1\ei{\begin{itemize}#1\end{itemize}}

\newcommand{\lefto}{\mathopen{}\left}

\newcommand{\vecnorm}[1]{\lefto\lVert#1\right\rVert}		

\safemath{\dirac}{\delta}					
\safemath{\krond}{\dirac}					

\safemath{\upto}{\uparrow}
\safemath{\downto}{\downarrow}
\safemath{\iu}{j}							
\safemath{\ev}{\lambda}						
\safemath{\hilseqspace}{l^{2}}				
\newcommand{\banachfunspace}[1]{\setL^{#1}}	
\safemath{\hilfunspace}{\banachfunspace{2}}	

\safemath{\SNR}{\text{\sc snr}} 				
\safemath{\No}{N_0}							
\safemath{\Es}{E_s}							
\safemath{\Eb}{E_b}							
\safemath{\EbNo}{\frac{\Eb}{\No}}
\safemath{\EsNo}{\frac{\Es}{\No}}

\DeclareMathOperator{\CHop}{\ensuremath{\opH}} 
\safemath{\tvir}{\rndh_{\CHop}}				
\safemath{\tvtf}{\rndl_{\CHop}}				
\safemath{\spf}{\rnds_{\CHop}}				
\safemath{\bff}{H_{\CHop}}					

\safemath{\ircf}{r_{h}}						
\safemath{\tftvcf}{r_{s}}					
\safemath{\tfcf}{r_{l}}						
\safemath{\bfcf}{r_{H}}						

\safemath{\tcorr}{c_h}						
\safemath{\scf}{c_{s}}						
\safemath{\tfcorr}{c_{l}}					
\safemath{\fcorr}{c_{H}}						

\safemath{\mi}{I}							
\safemath{\capacity}{C}						

\safemath{\normal}{\mathcal{N}}			
\safemath{\jpg}{\mathcal{CN}}			
\safemath{\mchain}{\leftrightarrow}		

\safemath{\dB}{\,\mathrm{dB}}
\safemath{\dBm}{\,\mathrm{dBm}}
\safemath{\Hz}{\,\mathrm{Hz}}
\safemath{\kHz}{\,\mathrm{kHz}}
\safemath{\MHz}{\,\mathrm{MHz}}
\safemath{\GHz}{\,\mathrm{GHz}}
\safemath{\s}{\,\mathrm{s}}
\safemath{\ms}{\,\mathrm{ms}}
\safemath{\mus}{\,\mathrm{\text{\textmu}s}}
\safemath{\ns}{\,\mathrm{ns}}
\safemath{\ps}{\,\mathrm{ps}}
\safemath{\meter}{\,\mathrm{m}}
\safemath{\mm}{\,\mathrm{mm}}
\safemath{\cm}{\,\mathrm{cm}}
\safemath{\m}{\,\mathrm{m}}
\safemath{\W}{\,\mathrm{W}}
\safemath{\mW}{\, \mathrm{mW}}
\safemath{\J}{\,\mathrm{J}}
\safemath{\K}{\,\mathrm{K}}
\safemath{\bit}{\,\mathrm{bit}}
\safemath{\nat}{\,\mathrm{nat}}

\safemath{\define}{\triangleq}			

\safemath{\equivalent}{\sim}
\safemath{\distas}{\sim}					
\safemath{\sdiff}{\Delta}				

\safemath{\reals}{\mathbb{R}}
\safemath{\positivereals}{\reals_{+}}
\safemath{\integers}{\mathbb{Z}}
\safemath{\posint}{\integers_{+}}
\safemath{\naturals}{\mathbb{N}}
\safemath{\posnaturals}{\naturals_{+}}
\safemath{\complexset}{\mathbb{C}}
\safemath{\rationals}{\mathbb{Q}}

\newcommand*{\fancyrefapplabelprefix}{app}		
\newcommand*{\fancyrefthmlabelprefix}{thm}		
\newcommand*{\fancyreflemlabelprefix}{lem}		
\newcommand*{\fancyrefcorlabelprefix}{cor}		
\newcommand*{\fancyrefdeflabelprefix}{def}		
\newcommand*{\fancyrefalglabelprefix}{alg}		
\newcommand*{\fancyrefproplabelprefix}{prop}		
\newcommand*{\fancyrefexmpllabelprefix}{exmpl}
\newcommand*{\fancyreftbllabelprefix}{tbl}
\frefformat{vario}{\fancyrefseclabelprefix}{Sec.~#1}
\frefformat{vario}{\fancyrefthmlabelprefix}{Thm.~#1}
\frefformat{vario}{\fancyreflemlabelprefix}{Lem.~#1}
\frefformat{vario}{\fancyrefcorlabelprefix}{Cor.~#1}
\frefformat{vario}{\fancyrefdeflabelprefix}{Def.~#1}
\frefformat{vario}{\fancyrefalglabelprefix}{Alg.~#1}
\frefformat{vario}{\fancyreffiglabelprefix}{Fig.~#1}
\frefformat{vario}{\fancyrefapplabelprefix}{App.~#1} 
\frefformat{vario}{\fancyrefeqlabelprefix}{(#1)}
\frefformat{vario}{\fancyrefproplabelprefix}{Prop.~#1}
\frefformat{vario}{\fancyrefexmpllabelprefix}{Ex.~#1}
\frefformat{vario}{\fancyreftbllabelprefix}{Tbl.~#1}

\safemath{\dictab}{[\,\dicta\,\,\dictb\,]}

\safemath{\ysig}{\bmy}
\safemath{\ysighat}{\hat{\ysig}}
\safemath{\ysigdim}{M}
\safemath{\xsig}{\bmx}
\safemath{\xsigdim}{N}
\safemath{\nx}{n_x}
\safemath{\zsig}{\bmz}
\safemath{\zsigdim}{\ysigdim}
\safemath{\rsig}{\bmr}
\safemath{\Adict}{\bA}
\safemath{\Adicttilde}{\widetilde{\Adict}}
\safemath{\Adictdim}{\outputdim\times\xsigdim}
\safemath{\avec}{\bma}
\safemath{\avectilde}{\tilde{\avec}}
\safemath{\Bdict}{\bB}
\safemath{\Bdicttilde}{\widetilde{\Bdict}}
\safemath{\Cdict}{\bC}
\safemath{\cvec}{\bmc}
\safemath{\Ddict}{\bD}
\safemath{\Ddictdim}{\ysigdim\times\xsigdim}
\safemath{\dvec}{\bmd}
\safemath{\Ddicttilde}{\widetilde{\bD}}
\safemath{\Bonb}{\bB}
\safemath{\bvec}{\bmb}
\safemath{\Bonbdim}{\ysigdim\times\ysigdim}
\safemath{\noise}{\bmn}
\safemath{\noisedim}{\ysigim}
\safemath{\err}{\bme}
\safemath{\errdim}{\ysigdim}
\safemath{\errset}{\setE}
\safemath{\nerr}{n_e}
\safemath{\delop}{\bP_\errset}
\safemath{\delopc}{\bP_{{\errset}^c}}

\safemath{\cplxi}{\imath}
\safemath{\cplxj}{\jmath}

\safemath{\dict}{\matD}
\safemath{\inputdim}{N}		
\safemath{\outputdim}{M}		
\safemath{\sparsity}{S}	
\safemath{\inputdimA}{{N_a}}	
\safemath{\inputdimB}{{N_b}}	
\safemath{\elemA}{{n_a}}	
\safemath{\elemB}{{n_b}}	
\safemath{\resA}{\matR_a}	
\safemath{\resB}{\matR_b}	
\safemath{\subD}{\matS} 
\safemath{\subA}{\matS_a} 
\safemath{\subB}{\matS_b} 
\safemath{\dicta}{\matA} 	
\safemath{\dictb}{\matB} 	
\safemath{\hollowS}{H}
\safemath{\hollowA}{H_a}
\safemath{\hollowB}{H_b}
\safemath{\cross}{Z}
\safemath{\coh}{\mu_d}			
\safemath{\coha}{\mu_a}			
\safemath{\cohb}{\mu_b}			
\safemath{\mubs}{\nu}	
\safemath{\cohm}{\mu_m} 
\safemath{\dictset}{\setD}	
\safemath{\dictsetp}{\dictset(\coh,\coha,\cohb)}	
\safemath{\dictsetgen}{\dictset_\text{gen}}
\safemath{\dictsetgenp}{\dictsetgen(\coh)}
\safemath{\dictsetonb}{\dictset_\text{onb}}
\safemath{\dictsetonbp}{\dictsetonb(\coh)}

\safemath{\leftside}{U}
\safemath{\rightsideA}{R_a}
\safemath{\rightsideB}{R_b}

\safemath{\indexS}{\setI_S} 

\safemath{\na}{n_a}			
\safemath{\nb}{n_b}			
\safemath{\coeffa}{p_i}	
\safemath{\coeffb}{q_j}	
\safemath{\seta}{\setP}		
\safemath{\setb}{\setQ}     
\safemath{\setw}{\setW}	
\safemath{\setz}{\setZ}	
\safemath{\cola}{\veca}		
\safemath{\colb}{\vecb}		
\safemath{\cold}{\vecd}		
\safemath{\inputvec}{\vecx} 	
\safemath{\error}{\vece}	
\safemath{\noiseout}{\vecz} 	
\safemath{\inputvecel}{x}
\safemath{\inputveca}{\vecx_a}
\safemath{\inputvecb}{\vecx_b}
\safemath{\outputvec}{\vecy}	
\safemath{\lambdamin}{\lambda_{\mathrm{min}}}

\newcommand{\normone}[1]{\vecnorm{#1}_1}

\safemath{\elltwo}{\ell_2}
\safemath{\ellone}{\ell_1}
\safemath{\ellzero}{\ell_0}
\safemath{\ellinf}{\ell_\infty}
\safemath{\licard}{Z(\coh,\coha,\cohb)}
\safemath{\xsol}{\hat{x}}
\safemath{\xbord}{x_b}		
\safemath{\xstat}{x_s}		
\safemath{\xstatLone}{\tilde{x}_s}
\safemath{\order}{\mathcal{O}} 
\safemath{\scales}{\Theta} 
\safemath{\ones}{\mathbf{1}} 
\safemath{\zeroes}{\mathbf{0}} 
\safemath{\thlone}{\kappa(\coh,\cohb)} 
\safemath{\constoneA}{\delta} 
\safemath{\constoneB}{\epsilon} 
\safemath{\nlarge}{L}				   
\safemath{\sumlarge}{S_\nlarge}
\safemath{\maxlarger}{P_\nlarge}	   
\safemath{\Pzero}{\textrm{P0}}	
\safemath{\Pone}{\textrm{P1}}
\safemath{\vecfir}{\vecw}			 
\safemath{\vecsec}{\vecz}
\safemath{\elvecfir}{w}              
\safemath{\elvecsec}{z}				 
\safemath{\nlargefir}{n}
\safemath{\normout}{\gamma}
\safemath{\auxfun}{h}
\safemath{\supp}{\textrm{supp}}

\safemath{\indexa}{\ell}
\safemath{\indexb}{r}
\safemath{\indexc}{i}
\safemath{\indexd}{j}

\safemath{\project}{P}

\begin{document}

\title{Tag-Aware Ordinal Sparse Factor Analysis \\[0.1cm] for Learning and Content Analytics}
\numberofauthors{1}
\author{
Andrew S. Lan, 
Christoph Studer, Andrew E. Waters,  
Richard G. Baraniuk\\[0.25cm]
       \affaddr{Rice University, TX, USA}\\[0.1cm]
       \email{\{mr.lan,\,studer,\,waters,\,richb\}@sparfa.com}}
\date{today}
\maketitle


%
            



\begin{abstract}

Machine learning offers novel ways and means to design {\em personalized learning systems} wherein each student's educational experience is customized in real time depending on their background, learning goals, and performance to date.
SPARse Factor Analysis (SPARFA) is a novel framework for machine learning-based {\em learning analytics}, which estimates a learner's knowledge of the concepts underlying a domain, and {\em content analytics}, which estimates the relationships among a collection of questions and those concepts.
SPARFA jointly learns the associations among the questions and the concepts, learner concept knowledge profiles, and the underlying question difficulties, solely based  on the correct/incorrect graded responses of a population of learners to a collection of questions.  
In this paper, we extend the SPARFA framework significantly to enable: (i) the analysis of graded responses on an ordinal scale (partial credit) rather than a binary scale (correct/incorrect); (ii) the exploitation of tags/labels for questions that partially describe the question--concept associations.  
The resulting {\em Ordinal SPARFA-Tag} framework greatly enhances the interpretability of the estimated concepts.
We demonstrate using real educational data that Ordinal SPARFA-Tag outperforms both SPARFA and existing collaborative filtering techniques in predicting missing learner responses.

\end{abstract}



\vspace{-0.2cm}

\section{Introduction} \label{sec:intro}

Today's education system typically provides only a ``one-size-fits-all'' learning experience that does not cater to the background, interests, and goals of individual learners. 
Modern machine learning (ML) techniques provide a golden opportunity to reinvent the way we teach and learn by making it more personalized and, hence, more efficient and effective.
The last decades have seen a great acceleration in the development of \emph{personalized learning systems} (PLSs), which can be grouped into two broad categories: (i) high-quality, but labor-intensive rule-based systems designed by domain experts that are hard-coded to give feedback in pre-defined scenarios \cite{pittcmu}, and (ii) more affordable and scalable ML-based systems that mine various forms of learner data in order to make performance predictions for each learner \cite{khan,huweb,edm}.

\vspace{-0.2cm}

\subsection{Learning and content analytics}
\sloppy
{\em Learning analytics} (LA, estimating what a learner understands based on data obtained from tracking their interactions with learning content) and {\em content analytics} (CA, organizing learning content such as questions, instructional text, and feedback hints) enable a PLS to generate automatic, targeted feedback to learners, their instructors, and content authors \cite{kulik1994meta}.
Recently we proposed a new framework for LA and CA based on SPARse Factor Analysis (SPARFA)~\cite{sparfa}. 
SPARFA consists of a statistical model and convex-optimization-based inference algorithms for analytics that leverage the fact that the knowledge in a given subject can typically be decomposed into a small set of latent knowledge components that we term \emph{concepts}~\cite{sparfa}. 
Leveraging the latent concepts and based only on the graded binary-valued  responses (i.e., correct/incorrect) to a set of questions, SPARFA jointly estimates (i) the associations among the questions and the concepts (via a ``concept graph''), (ii) learner concept knowledge profiles, and (iii) the underlying question difficulties.  
\fussy

\sloppy

\subsection{Contributions}
In this paper, we develop {\em Ordinal SPARFA-Tag}, a significant extension to the SPARFA framework that enables the exploitation of the additional information that is often available in educational settings. 
First, Ordinal SPARFA-Tag exploits the fact that responses are often graded on an ordinal scale (partial credit), rather than on a binary scale (correct/incorrect). 
Second, Ordinal SPARFA-Tag exploits tags/labels (i.e., keywords characterizing the underlying knowledge component related to a question) that can be attached by instructors and other users to questions. 
Exploiting pre-specified tags within the estimation procedure provides significantly more interpretable question--concept associations. 
Furthermore, our statistical framework can discover new concept--question relationships that would not be in the pre-specified tag information but, nonetheless, explain the graded learner--response data.

We showcase the superiority of  Ordinal SPARFA-Tag compared to the methods in \cite{sparfa} via a set of synthetic ``ground truth'' simulations and on a variety of experiments with real-world educational datasets. 
We also demonstrate that Ordinal SPARFA-Tag outperforms existing state-of-the-art collaborative filtering techniques in terms of predicting missing ordinal learner responses.
\fussy

\section{Statistical Model} 
\label{sec:model}

\sloppy

We assume that the learners' knowledge level on a set of abstract latent \emph{concepts} govern the responses they provide to a set of questions.
The SPARFA statistical model characterizes the probability of learners' binary (correct/incorrect) graded responses to questions in terms of three factors: (i) question--concept associations, (ii) learners' concept knowledge, and (iii) intrinsic question  difficulties; details can be found in \cite[Sec.~2]{sparfa}.
In this section, we will first extend the SPARFA framework to characterize \emph{ordinal} (rather than binary-valued) responses, and then impose additional structure in order to model real-world educational behavior more accurately.

\fussy

\subsection{Model for ordinal learner response data}
\label{sec:omodel}

Suppose that we have $N$ learners, $Q$ questions, and $K$ underlying concepts. 
Let $Y_{i,j}$ represent the graded response (i.e., score) of the $j^\text{th}$ learner to the $i^\text{th}$ question, which are from a set of $P$ ordered labels, i.e., $Y_{i,j} \in \mathcal{O}$, where $\mathcal{O} = \{1, \ldots, P\} $. 
For the $i^\text{th}$ question, with $i \in \{1,\ldots,Q\}$, we propose the following model for the learner--response relationships: 
\begin{align} 
 \label{eq:qa} & Z_{i,j} = \vecw_i^T \vecc_j + \mu_i, \,\, \forall (i,j), \quad  \\  & Y_{i,j} = \mathcal{Q}(Z_{i,j}+\epsilon_{i,j}),  \; \epsilon_{i,j} \sim \mathcal{N}\!\left(0,{1}/{\tau_{i,j}}\right), \; (i,j)\in\Omega_\text{obs},  \notag
\end{align}
where the column vector $\vecw_i \in \mathbb{R}^{K}$ models the \emph{concept associations}; i.e., it encodes how question $i$ is related to each concept. Let the column vector $\bmc_j \in \mathbb{R}^{K}$, $j \in \{1,\ldots,N\}$, represent the latent \emph{concept knowledge} of the $j^\text{th}$ learner, with its $k^{\text{th}}$ component representing the $j^\text{th}$ learner's knowledge of the  $k^\text{th}$ concept. The scalar $\mu_i$ models the \emph{intrinsic difficulty} of question~$i$, with large positive value of $\mu$ for an easy question. 
The quantity $\epsilon_{i,j}$ models the uncertainty of learner $j$ answering question $i$ correctly/incorrectly and 
$\mathcal{N}(0,1/\tau_{i,j})$ denotes a zero-mean Gaussian distribution with precision parameter $\tau_{i,j}$, which models the \emph{reliability} of the observation of learner $j$ answering question $i$. 
We will further assume $\tau_{i,j} = \tau$, meaning that all the observations have the same reliability.\footnote{Accounting for learner/question-varying reliabilities is straightforward and omitted for the sake of brevity.}
The slack variable $Z_{i,j}$ in~\fref{eq:qa} governs the probability of the observed grade $Y_{i,j}$. The set $\Omega_\text{obs}\subseteq\{1,\ldots,Q\}\times\{1,\ldots,N\}$ contains the indices associated to the observed learner--response data, in case the response data is not fully observed.

In \fref{eq:qa}, $\mathcal{Q}(\cdot): \mathbb{R} \rightarrow \mathcal{O}$ is a scalar quantizer that maps a real number into $P$ ordered labels according to 
\begin{align*}
\mathcal{Q}(x) = 
p \quad \text{if }\, \omega_{p-1}< x \leq \omega_p,\,\,   p \in \setO,
\end{align*}
where $\{\omega_0, \ldots, \omega_P \}$ is the set of quantization bin boundaries satisfying $\omega_0 < \omega_1 < \cdots < \omega_{P-1} < \omega_P$, with $\omega_0$ and $\omega_P$ denoting the lower and upper bound of the domain of the quantizer $\mathcal{Q}(\cdot)$.\footnote{In most situations, we have $\omega_0 = -\infty$ and $\omega_P = \infty$.} 
This quantization model leads to the equivalent input--output relation
\begin{align}
\label{eq:qap} & Z_{i,j} = \vecw_i^T \vecc_j + \mu_i,  \quad \forall (i,j), \quad  \text{and} \;\; \\ \notag &  p(Y_{i,j}=p \mid Z_{i,j}) = \int_{\omega_{p-1}}^{\omega_p} \mathcal{N}(s | Z_{i,j},1/\tau_{i,j}) \, \mathrm{d}s   \\ \notag
 &\quad \! = \, \Phi \! \left( \tau( \omega_p \! - \! Z_{i,j}) \right) \! - \! \Phi \! \left( \tau(\omega_{p-1} \! - \! Z_{i,j})\right),  (i,j) \! \in \! \Omega_\text{obs},  
\end{align}
where $\Phi(x) = \int_{-\infty}^x \! \mathcal{N}(s | 0,1)  \mathrm{d}s$ denotes the \emph{inverse probit function}, with $\setN(s | 0,1)$ representing the value of a standard normal distribution evaluated at $s$.\footnote{Space limitations preclude us from discussing a corresponding logistic-based model; the extension is straightforward.} 
We can conveniently rewrite \fref{eq:qa} and \fref{eq:qap} in matrix form as
\begin{align} \label{eq:qam}
 & \bZ = \bW \bC, \quad \forall (i,j), \quad \text{and} \notag \\
 & p(Y_{i,j} \mid Z_{i,j})  = \Phi \! \left( \tau(U_{i,j} - Z_{i,j}) \right) \\
\notag & \qquad  \,\,\, \qquad \qquad  - \Phi \! \left( \tau(L_{i,j} - Z_{i,j}) \right) ,\,\, (i,j) \in \Omega_\text{obs},
\end{align}
where $\bY$ and $\bZ$ are $Q \times N$ matrices. The $Q \times (K+1)$ matrix $\bW$ is  formed by concatenating $[\vecw_1, \ldots, \vecw_Q]^T$ with the intrinsic difficulty vector $\boldsymbol{\mu}$ and $\bC$ is a $(K+1) \times N$ matrix formed by concatenating the $K \times N$ matrix $[\vecc_1, \ldots, \vecc_N]$ with an all-ones row vector $\bOne_{1\times N}$. 
We furthermore define the \mbox{$Q \times N$} matrices $\bU$ and $\bL$ to contain the upper and lower bin  boundaries corresponding to the observations in $\bY$, i.e., we have $U_{i,j} \!=\! \omega_{Y_{i,j}}$ and $L_{i,j} \! = \!\omega_{Y_{i,j}-1}$, $\forall (i,j) \in \Omega_\text{obs}$.

We emphasize that the statistical model proposed above is significantly more general than the original SPARFA model proposed in \cite{sparfa}, which is a special case of \fref{eq:qa} with $P=2$ and $\tau = 1$. The precision parameter $\tau$ does not play a central role  in \cite{sparfa} (it has been set to $\tau=1$), since the observations are  binary-valued with bin boundaries $\{ -\infty, 0, \infty\}$. For ordinal responses (with $P>2$), however, the precision parameter~$\tau$ significantly affects the behavior of the statistical model and, hence, we estimate the precision parameter~$\tau$ directly from the observed data.

\subsection{Fundamental assumptions}
\label{sec:assumptions}

\sloppy

Estimating $\bW$, $\boldsymbol{\mu}$ and $\bC$ from $\bY$ is an ill-posed problem, in general, since there are more unknowns than observations and the observations are ordinal (and not real-valued). To ameliorate the illposedness, \cite{sparfa} proposed three assumptions accounting for real-world educational situations: 
\vspace*{-5mm}
\begin{enumerate}
\item[(A1)] \emph{Low-dimensionality}: Redundancy exists among the questions in an assessment, and the observed graded learner responses live in a low-dimensional space, i.e., \mbox{$K \ll N$},~$Q$. 
\vspace*{-1mm}
\item[(A2)] \emph{Sparsity}: Each question measures the learners' knowledge on only a few concepts (relative to $N$ and $Q$), i.e., the question--concept association matrix $\bW$ is sparse. 
\vspace*{-5mm}
\item[(A3)] \emph{Non-negativity}: The learners' knowledge on concepts does not reduce the chance of receiving good score on any question, i.e., the entries in~$\bW$ are non-negative. Therefore, large positive values of the entries in $\bC$ represent good concept knowledge, and vice versa.
\end{enumerate}
\vspace*{-5mm}
Although these assumptions are reasonable for a wide range of educational contexts (see \cite{sparfa} for a detailed discussion), they are hardly complete. In particular, additional information is often available regarding the questions and the learners in some situations. Hence, we impose one additional assumption:
\vspace*{-5mm}
\begin{itemize}
\item[(A4)] \emph{Oracle support}: Instructor-provided tags on questions provide prior information on some question--concept associations. In particular, associating each tag with a {\em single concept} will partially (or fully) determine the locations of the non-zero entries in $\bW$. 
\end{itemize}
\vspace{-5mm}
As we will see, assumption (A4) significantly improves the limited interpretability of the estimated factors $\bW$ and $\bC$ over the conventional SPARFA framework~\cite{sparfa}, which relies on a (somewhat ad-hoc) post-processing step to associate instructor provided tags with concepts.
In contrast, we utilize the tags as ``oracle'' support information on~$\bW$ {\em within} the model, which enhances the explanatory performance of the statistical framework, i.e., it enables to associate each concept directly with a predefined tag.
Note that user-specified tags might not be precise or complete. Hence, the proposed estimation algorithm must be capable of discovering new question--concept associations and removing predefined associations that cannot be explained from the observed data.
%

\section{Algorithm} 
\label{sec:matrix}

We start by developing \emph{Ordinal SPARFA-M}, a generalization of SPARFA-M from \cite{sparfa} to ordinal response data. Then, we detail \emph{Ordinal SPARFA-Tag}, which considers prespecified question tags as oracle support information of~$\bW$, to estimate $\bW$, $\bC$, and $\tau$, from the ordinal response matrix $\bY$ while enforcing the assumptions (A1)--(A4).

\subsection{Ordinal SPARFA-M} 
\label{sec:outi}

To estimate $\bW$, $\bC$, and $\tau$ in \fref{eq:qam} given $\bY$, we maximize the log-likelihood of $\bY$ subject to (A1)--(A4)
by solving
\begin{align*}
(\text{P})  \quad  & \underset{\bW,\bC,\tau }{\text{minimize}} \textstyle - \! \sum_{i,j: (i,j) \in \Omega_\text{obs}} \log p(Y_{i,j}| \tau\vecw_i ^T \vecc_j) \\
&   \textstyle+ \lambda \sum_{i} \| \vecw_i \|_1 \,\,\,\, \text{subject to} \,\,\,\, \bW \geq 0, \tau>0, \|\bC\| \leq \eta.
\end{align*}
Here, the likelihood of each response $p(Y_{i,j}|\tau\vecw_i ^T \vecc_j)$ is given by \fref{eq:qap}. The regularization term $\lambda\sum_i\normone{\vecw_i}$ imposes sparsity on each vector $\vecw_i$ to account for~(A2).
To prevent arbitrary scaling between $\bW$ and $\bC$, we gauge the norm of the matrix~$\bC$ by applying a matrix norm constraint $\|\bC\| \leq \eta$. For example, the Frobenius norm constraint \mbox{$\|\bC\|_F \leq \eta$} can be used. Alternatively, the nuclear norm constraint \mbox{$\|\bC\|_* \leq \eta$} can also be used, promoting low-rankness of $\bC$~\cite{candessvt}, motivated by the facts that (i) reducing the number of degrees-of-freedom in $\bC$ helps to prevent  overfitting to the observed data and (ii) learners can often be clustered into a few groups due to their different demographic backgrounds and learning preferences.

The log-likelihood of the observations in (P) is concave in the product $\tau\vecw_i ^T \vecc_j$ \cite{candes}. 
Consequently, the problem~$(\text{P})$ is \emph{tri-convex}, in the sense that the problem obtained by holding two of the three factors $\bW, \bC$, and $\tau$ constant and optimizing the third one is convex. 
Therefore, to arrive at a practicable way of solving $(\text{P})$, we propose the following computationally efficient block coordinate descent approach, with $\bW$, $\bC$, and $\tau$ as the different blocks of variables.

The matrices $\bW$ and $\bC$ are initialized as i.i.d.\ standard normal random variables, and we set $\tau=1$.
We then iteratively optimize the objective of $(\text{P})$ for all three factors in round-robin fashion.
Each (outer) iteration consists of three phases: first, we hold~$\bW$ and $\tau$ constant and optimize $\bC$; second, we hold $\bC$ and~$\tau$ constant and separately optimize each row vector $\vecw_i$; third, we hold $\bW$ and $\bC$ fixed and optimize over the precision parameter $\tau$. These three phases form the outer loop of Ordinal SPARFA-M.

The sub-problems for estimating $\bW$ and~$\bC$ correspond to the following {\em ordinal regression} (OR) problems \cite{ordreg}: 
\begin{align*}
&(\text{OR-W})  \quad
\underset{\vecw_i\colon\!W_{i,k}\geq0\;\forall k}{\text{minimize}}\! \textstyle -  \!\sum\nolimits_{j} \log p(Y_{i,j}|\tau\vecw_i^T\vecc_j) \! + \! \lambda\normone{\vecw_i} ,  \\
&(\text{OR-C}) \,\,\quad \,\,\,\,
\underset{\bC: \|\bC\| \leq \eta}{\text{minimize}}\;\;\,\! \textstyle - \!\sum\nolimits_{i,j} \log p(Y_{i,j}|\tau\vecw_i^T\vecc_j).
\end{align*}
To solve (OR-W) and (OR-C), we deploy the iterative first-order methods detailed below.
To optimize the precision parameter $\tau$, we compute the solution to 
\begin{align*}
& \underset{\tau>0}{\text{minimize}}\; \textstyle -  \sum_{i,j: (i,j) \in \Omega_\text{obs}} \\[-0.05cm]
& \quad \log \!\left( \Phi \left(\tau\left(\bU_{i,j}-\vecw_i^T \vecc_j\right)\right) - \Phi \left(\tau\left(\bL_{i,j}-\vecw_i^T \vecc_j\right)\right)\right),
\end{align*}
via the secant method \cite{sjwright}.

Instead of fixing the quantization bin boundaries $\{\omega_0, \ldots, \omega_P \}$ introduced in \fref{sec:model} and optimizing the precision and intrinsic difficulty parameters, one can fix $\tau = 1$ and optimize the bin boundaries instead, an approach used in, e.g.,  \cite{ordpred}. 
We emphasize that optimization of the bin boundaries can also be performed straightforwardly via the secant method, iteratively optimizing each bin boundary while keeping the others fixed. We omit the details for the sake of brevity.
Note that we have also implemented variants of Ordinal SPARFA-M that directly optimize the bin boundaries, while keeping $\tau$ constant; the associated prediction performance is shown in \fref{sec:pred}.
\begin{figure*}[t]
\vspace{-1.0cm}
\centering

\subfigure[Impact of the number of learners, $N \in \{50,100,200\}$, with the number of questions $Q$ fixed.
\label{fig:varyn}]{\hspace{-0.2cm}
\includegraphics[width=0.32\textwidth]{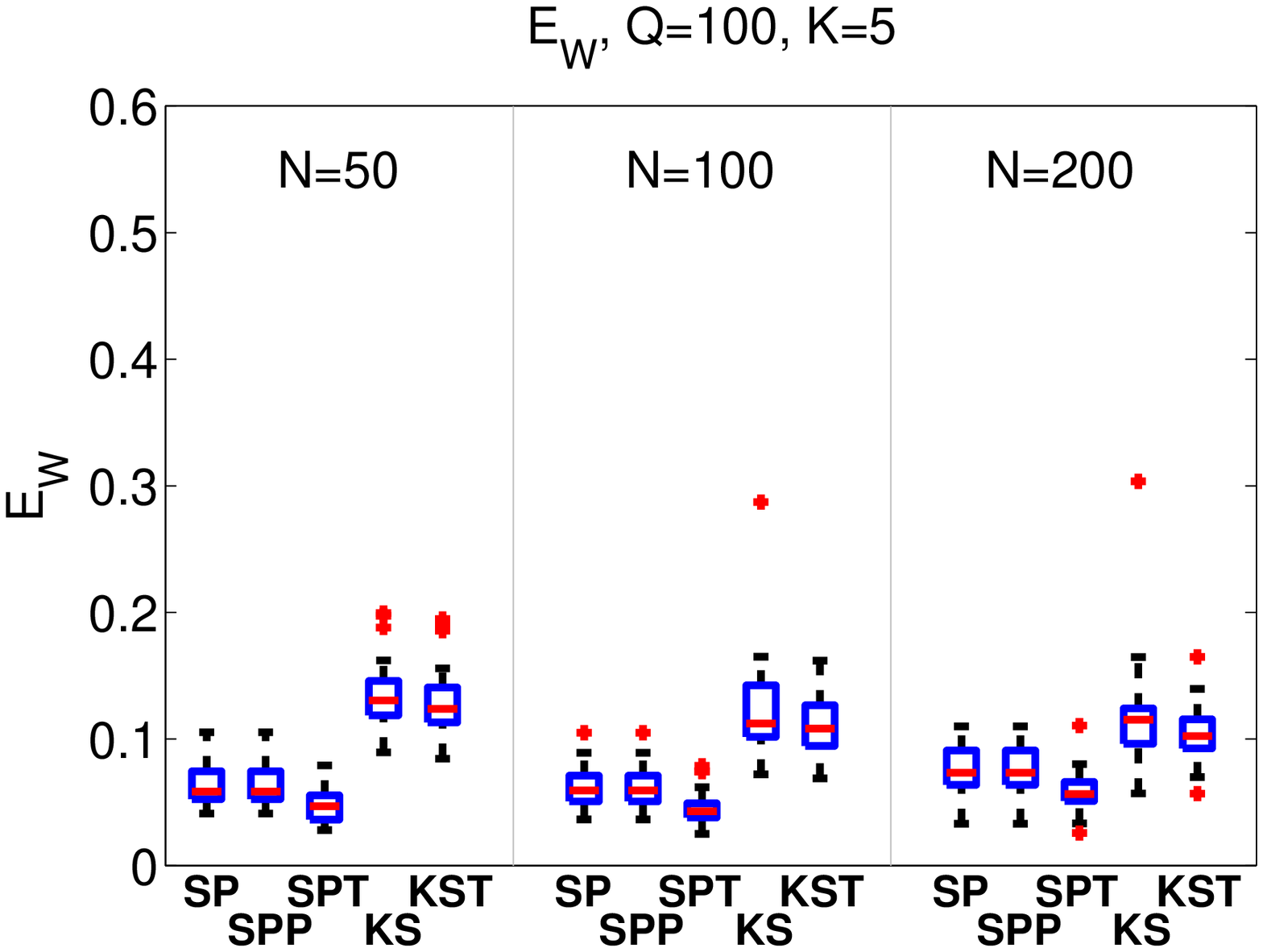} \hspace{-0.1cm}
\includegraphics[width=0.32\textwidth]{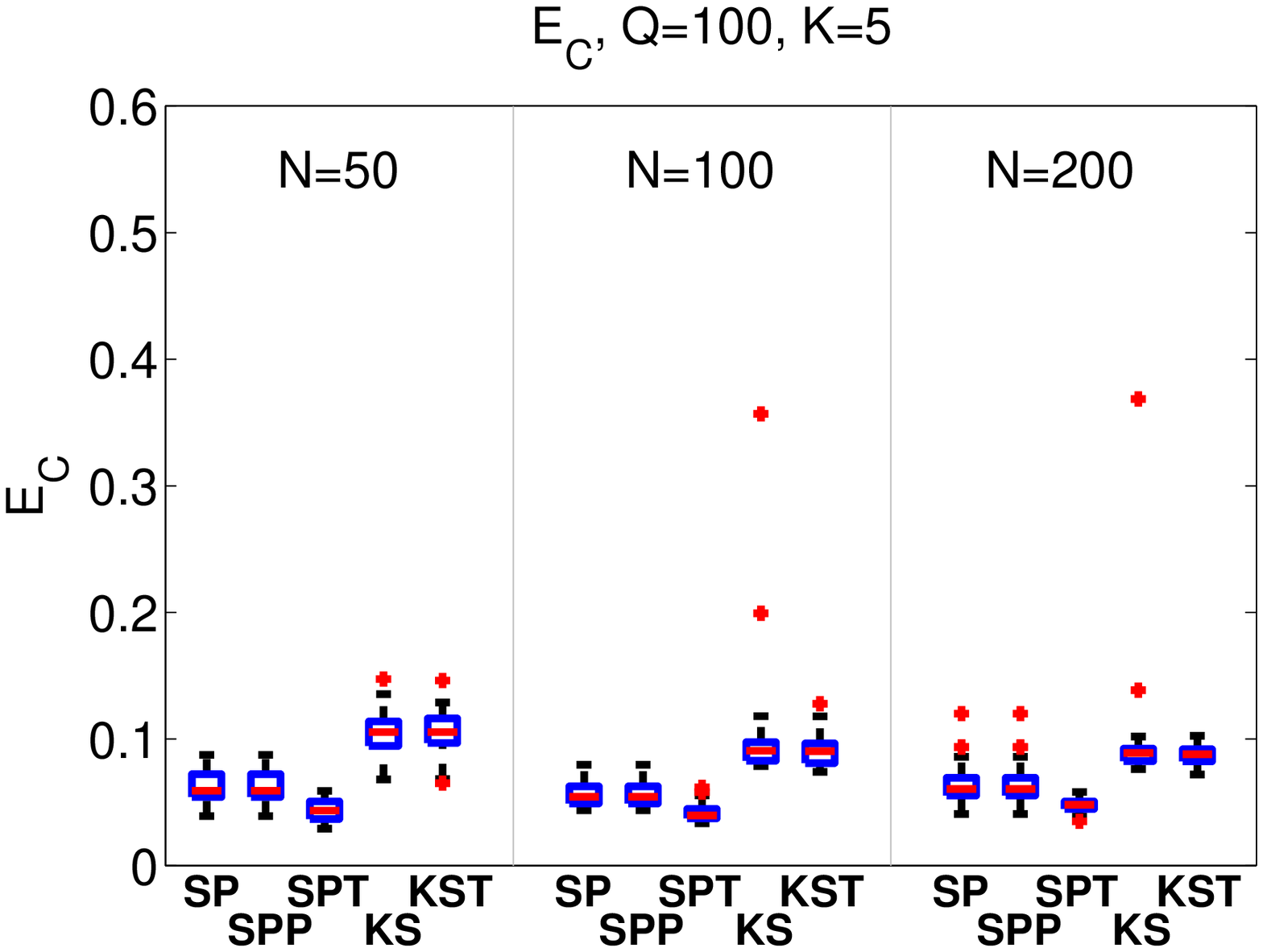} \hspace{-0.1cm}
\includegraphics[width=0.32\textwidth]{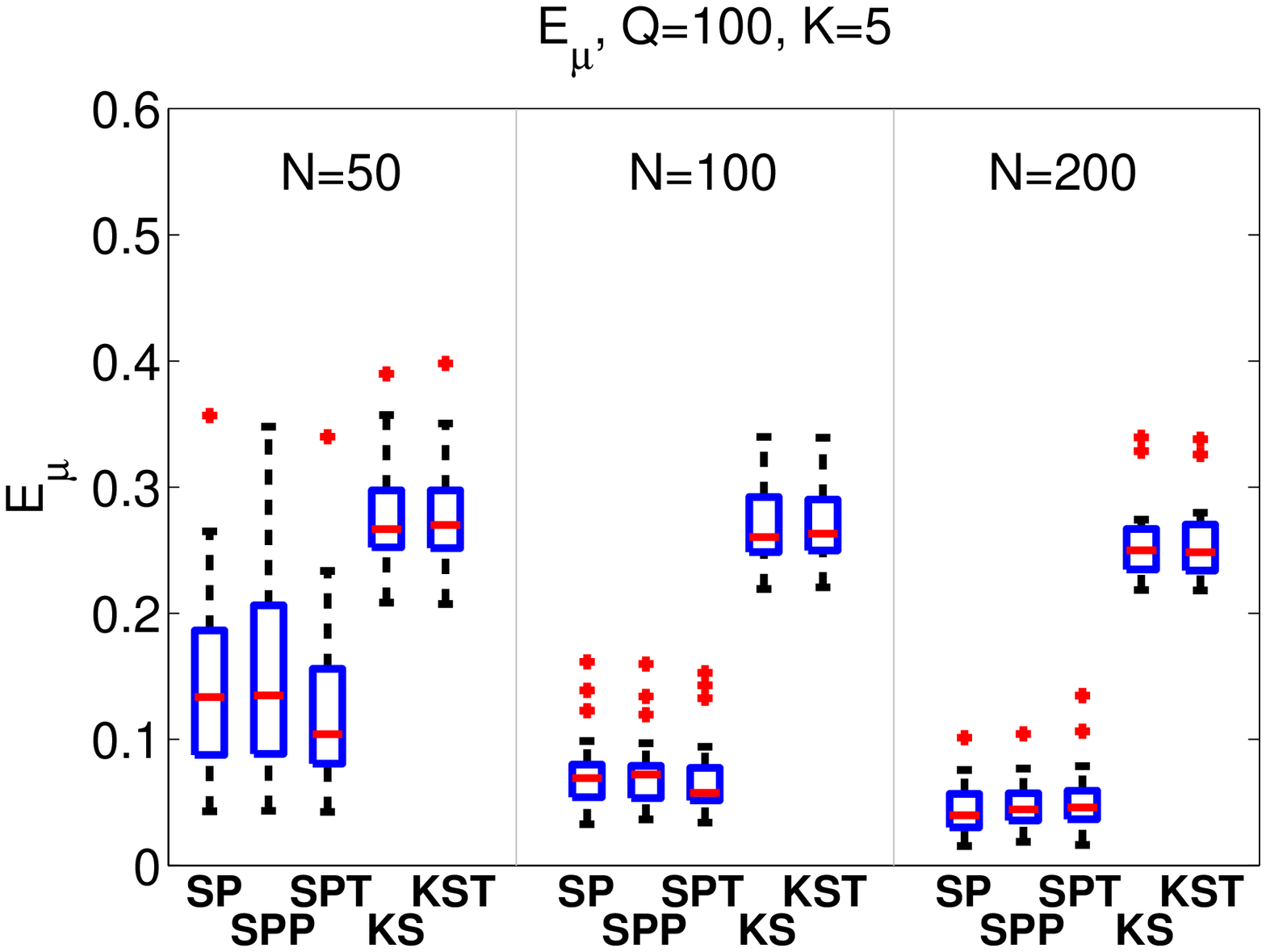}
\hspace{-0.0cm}}
\vspace{-0.2cm}
\subfigure[Impact of the number of questions, $Q \in \{50,100,200\}$, with the number of learners $N$ fixed.\label{fig:varyq}]{
\includegraphics[width=0.32\textwidth]{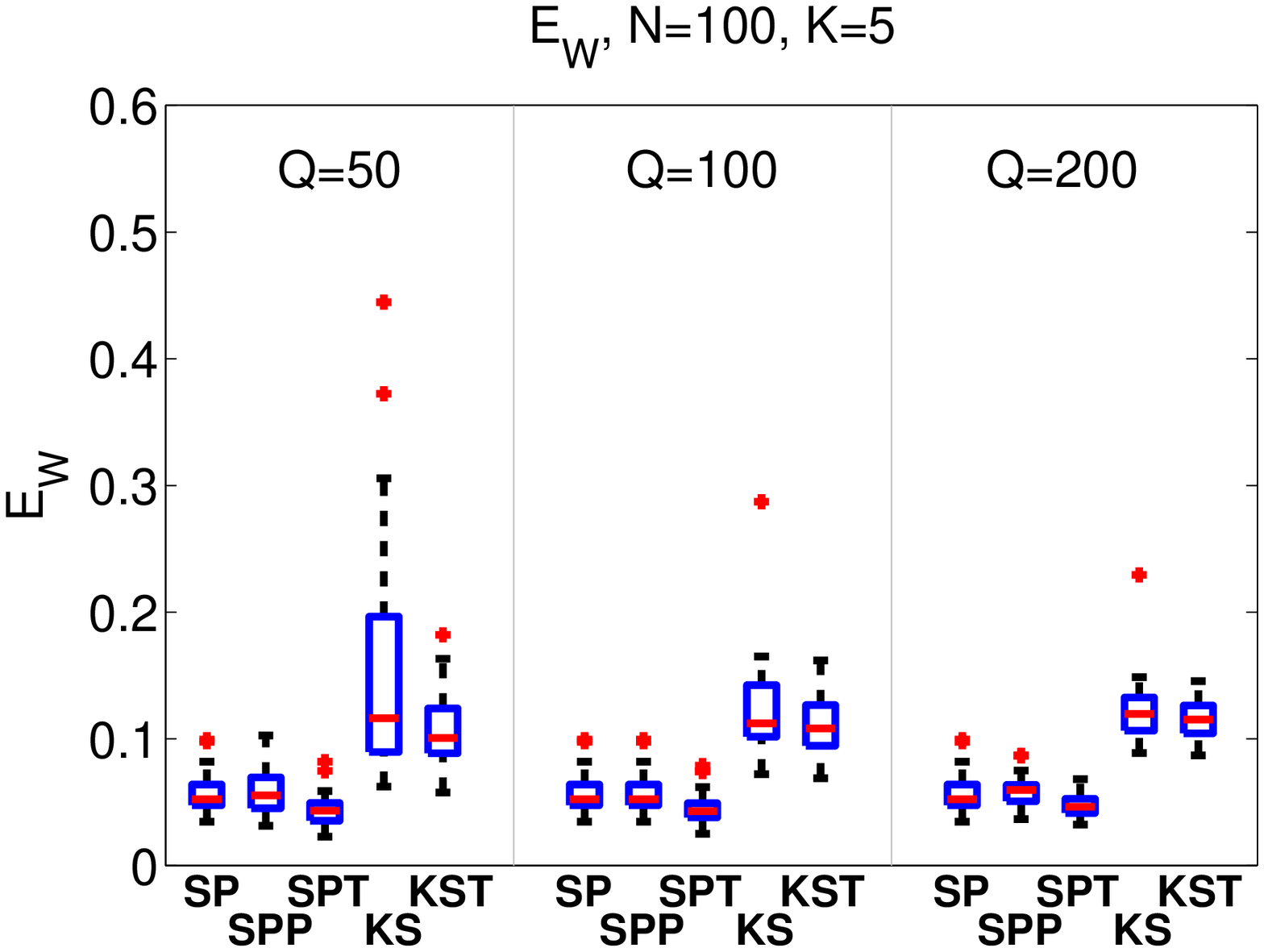} \hspace{-0.1cm}
\includegraphics[width=0.32\textwidth]{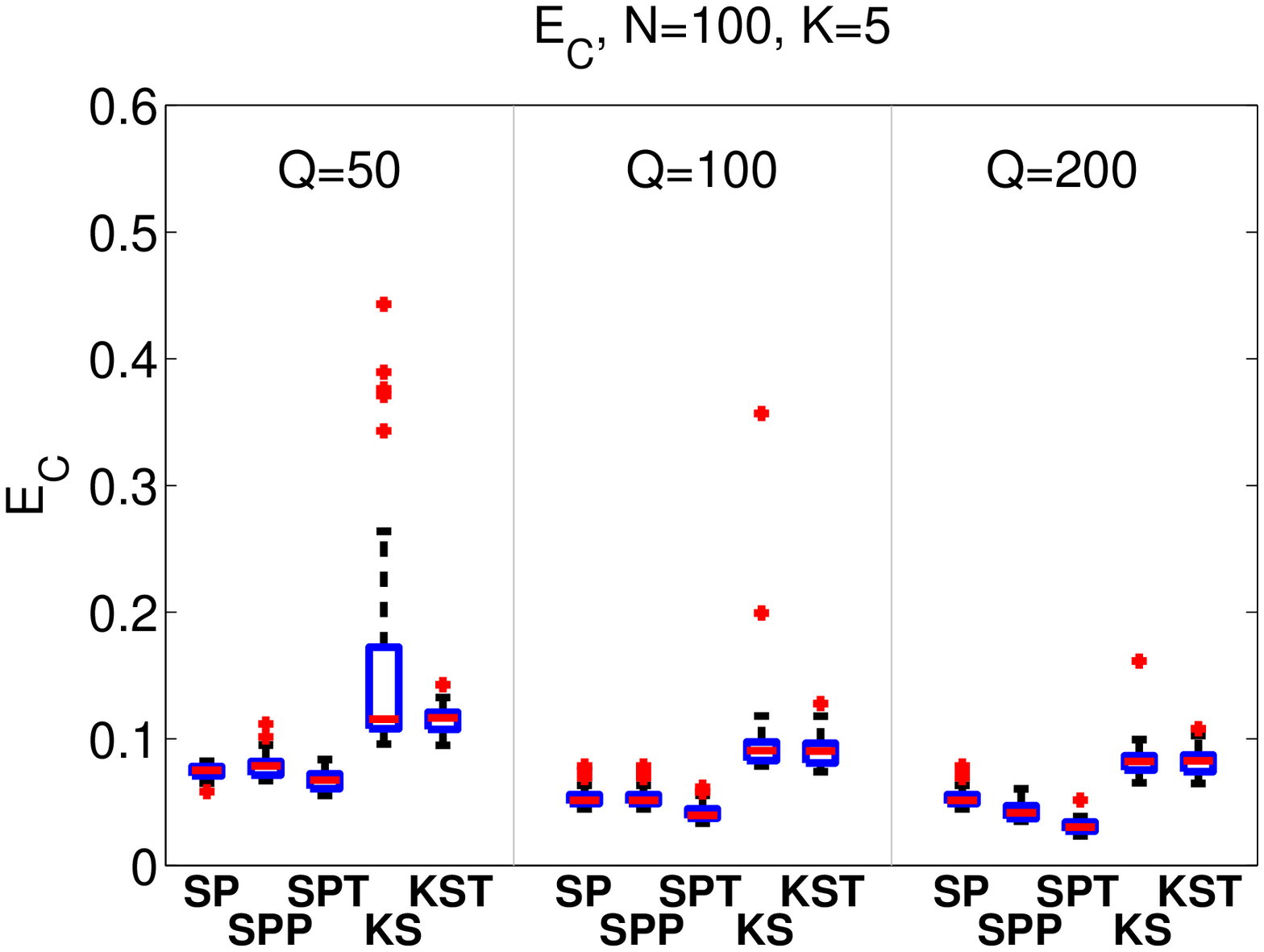} \hspace{-0.1cm}
\includegraphics[width=0.32\textwidth]{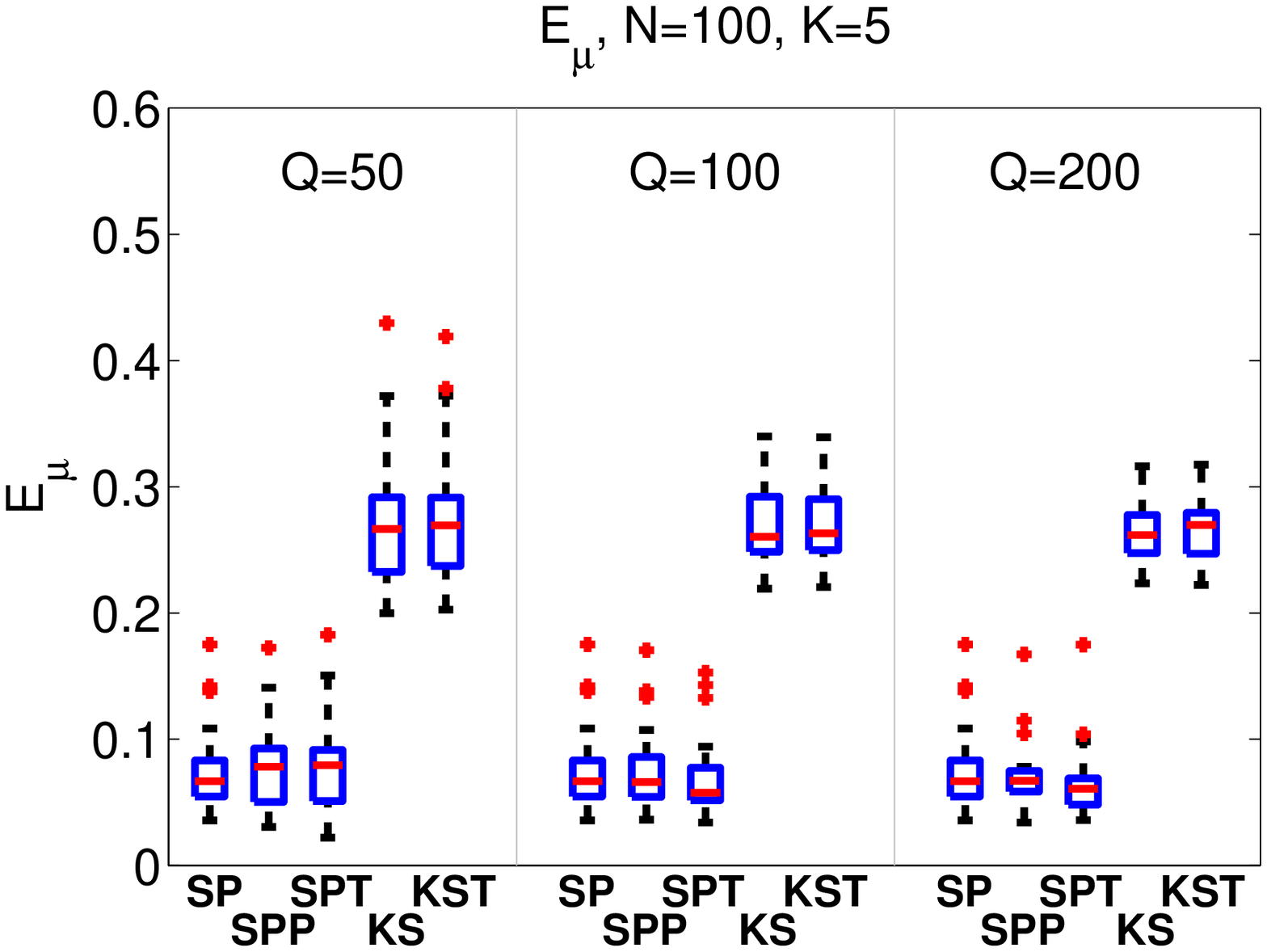}
\hspace{-0.0cm}}
\vspace{-0.1cm}
\caption{Performance comparison of Ordinal SPARFA-M vs. K-SVD$_+$. ``SP'' denotes Ordinal SPARFA-M without given support $\Gamma$ of $\bW$, ``SPP'' denotes the variant with estimated precision $\tau$, and ``SPT'' denotes Ordinal SPARFA-Tag. ``KS'' stands for K-SVD$_+$, and ``KST'' denotes its variant with given support $\Gamma$.}
\vspace{-0.3cm}
\label{fig:synthplots}
\vspace{-0.0cm}
\end{figure*}

\subsection{First-order methods for regularized \\ ordinal regression} 
\label{sec:RPR}

As in \cite{sparfa}, we solve (OR-W) using the FISTA framework~\cite{fista}.
(OR-C) also falls into the FISTA framework, by re-writing the convex constraint $\|\bC\| \leq \eta$ as a penalty term $\delta (\bC: \|\bC\| > \eta)$ and treat it as a non-smooth regularizer, where $\delta (\bC: \|\bC\| > \eta)$ is the delta function, equaling 0 if \mbox{$\|\bC\| \leq \eta$} and $\infty$ otherwise.
Each iteration of both algorithms consists of two steps: A gradient-descent step and a shrinkage/projection step. Take (OR-W), for example, and let $f(\vecw_i) = -\!\sum_{j} \log p(Y_{i,j}|\tau\vecw_i^T\vecc_j)$. Then, the gradient step is given by\footnote{Here, we assume $\Omega_\text{obs} = \{1, \ldots, Q\} \times \{1, \ldots, N\}$ for simplicity; a generalization to the case of missing entries in $\bY$ is straightforward.}%
\begin{align} \label{eq:grad}
\nabla f &=  \nabla_{\!\vecw_i} ( -\textstyle\sum\nolimits_{j} \log p(Y_{i,j}|\tau\vecw_i^T \vecc_j) ) =  -\bC \vecp.
\end{align}
Here, $\vecp$ is a $N \times 1$ vector, with the $j^\text{th}$ element equal to
\begin{align*}
\frac{\mathcal{N} (\tau (U_{i,j} - \vecw_i^T \vecc_j))-\mathcal{N} (\tau (L_{i,j} - \vecw_i^T \vecc_j))}{\Phi (\tau (U_{i,j} - \vecw_i^T \vecc_j)-\Phi (\tau (L_{i,j} - \vecw_i^T \vecc_j))},
\end{align*}
where $\Phi(\cdot)$ is the inverse probit function. 
The gradient step and the shrinkage step for $\vecw_i$ corresponds to
\begin{align} \label{eq:fistagradient}
\hat{\vecw}_i^{\ell+1} \gets \vecw_i^\ell - t_\ell \nabla f,
\end{align}
and
\begin{align}\label{eq:shrink1} 
\vecw_i^{\ell+1} \gets \max\{ \hat{\vecw}_i^{\ell+1}-\lambda t_\ell,0\},
\end{align}%
respectively, where $t_\ell$ is a suitable step-size.
For (OR-C), the gradient with respect to each column $\vecc_j$ is given by substituting $\bW^T$ for $\bC$ and $\vecc_j$ for $\vecw_i$ in~\fref{eq:grad}. Then, the gradient for $\bC$ is formed by aggregating all these individual gradient vectors for $\vecc_j$ into a corresponding gradient matrix. 

For the Frobenius norm constraint $\|\bC\|_F \leq \eta$, the projection step is given by \cite{boydbook}
\begin{align} \label{eq:shrink2} 
 \bC^{\ell+1} \gets \left \{ \begin{array}{ll}
\hat{\bC}^{\ell+1}  &\text{if} \,\, \|\hat{\bC}^{\ell+1}\|_F \leq \eta \\
\eta \frac{\hat{\bC}^{\ell+1}}{\|\hat{\bC}^{\ell+1}\|_F} & \text{otherwise}.
\end{array} \right.
\end{align} 
For the nuclear-norm constraint $\|\bC\|_* \leq \eta$, the projection step is given by
\begin{align} \label{eq:shrinknuc} 
& \bC^{\ell+1} \gets  \, \bU \text{diag}(\vecs) \bV^T, \,\text{with}\,\, \vecs = \text{Proj}_\eta(\text{diag}(\bS)),
\end{align} 
where $\hat{\bC}^{\ell+1} = \bU \bS \bV^T$ denotes the singular value decomposition, and $\text{Proj}_\eta(\cdot)$ is the projection onto the $\ellone$-ball with radius $\eta$ (see, e.g.,~\cite{l1proj} for the details).

The update steps~\fref{eq:fistagradient}, \fref{eq:shrink1}, and \fref{eq:shrink2} (or \fref{eq:shrinknuc}) require a suitable step-size $t_\ell$ to ensure convergence. We consider a constant step-size and set~$t_\ell$  to the reciprocal of the Lipschitz constant~\cite{fista}. 
The Lipschitz constants correspond to $\tau^2 \sigma^2_\text{max}(\bC)$ for (OR-W) and $\tau^2 \sigma^2_\text{max}(\bW)$ for (OR-C), with ${\sigma_\text{max}}(\bX)$ representing the maximum singular value of $\bX$.

\subsection{Ordinal SPARFA-Tag}
\label{sec:oracle}

We now develop the Ordinal SPARFA-Tag algorithm that incorporates (A4). 
Assume that the total number of tags associated with the $Q$ questions equal $K$ (each of the $K$ concepts correspond to a tag), and
define the set $\Gamma = \{(i,k): \text{question $i$ has tag $k$}\}$ as the set of indices of entries in $\bW$ identified by pre-defined tags, and $\bar{\Gamma}$ as the set of indices not in $\Gamma$, we can re-write the optimization problem (P) as: 
\begin{align*}
(\text{P}_\Gamma )  \quad \underset{\bW,\bC,\tau}{\text{minimize}}\,\,  &  \,\, \textstyle -  \sum_{i,j: (i,j) \in \Omega_\text{obs}} \log p(Y_{i,j}|\tau\vecw_i^T \vecc_j) \\[-0.2cm]
& \,\,+ \lambda \textstyle \sum_{i} \| \vecw_i^{(\bar{\Gamma})} \|_1 +  \gamma \sum_{i} \frac{1}{2}\| \vecw_i^{(\Gamma)} \|_2^2\\[0.1cm]
\text{subject to} &\,\,\, \bW \geq 0, \tau>0,\|\bC\| \leq \eta.
\end{align*}
Here, $\vecw_i^{(\Gamma)}$ is a vector of those entries in $\vecw_i$ belonging to the set $\Gamma$, while $\vecw_i^{(\bar{\Gamma})}$ is a vector of entries in $\vecw_i$ not belonging to~$\Gamma$. 
The $\elltwo$-penalty term on $\vecw_i^{(\Gamma)}$ regularizes the entries in $\bW$ that are part of the (predefined) support of $\bW$; we set \mbox{$\gamma=10^{-6}$ in all our experiments}.
The $\ellone$-penalty term on $\vecw_i^{(\bar{\Gamma})}$ induces sparsity on the entries in $\bW$ that are not predefined but might be in the support of $\bW$. 
Reducing the parameter $\lambda$ enables one to discover new question--concept relationships (corresponding to new non-zero entries in~$\bW$) that were not contained in $\Gamma$.

The problem $(\text{P}_\Gamma )$ is solved analogously to the approach described in~\fref{sec:RPR}, except that we split the $\bW$ update step into two parts that operate separately on the entries indexed by $\Gamma$ and $\bar{\Gamma}$. For the entries in $\Gamma$, the projection step corresponds to 
\begin{align*}
\vecw_i^{(\Gamma),\ell+1} \gets \text{max} \{\hat{\vecw}_i^{(\Gamma),\ell+1}/(1+\gamma t_\ell), 0\}. 
\end{align*}
The step for the entries indexed by $\bar{\Gamma}$ is given by \fref{eq:shrink1}.
Since Ordinal SPARFA-Tag is tri-convex, it does not necessarily converge to a global optimum. Nevertheless, we can leverage recent results in \cite{sparfa,wotao} in order to show that Ordinal SPARFA-Tag converges to a local optimum from an arbitrary starting point.
Furthermore, if the starting point is within a close neighborhood of a global optimum of (P), then Ordinal SPARFA-Tag converges to this global optimum.



\section{Experiments}
\label{sec:experiments}

\sloppy

We first showcase the performance of Ordinal SPARFA-Tag on synthetic data to demonstrate its convergence to a known ground truth. We then demonstrate the ease of interpretation of the estimated factors by leveraging instructor provided tags in combination with a Frobenius or nuclear norm constraint for two real educational datasets. We finally compare the performance of Ordinal SPARFA-M to state-of-the-art collaborative filtering techniques on predicting unobserved ordinal learner responses.

\fussy

\begin{figure*}[t]
\vspace{-.5cm}
\centering
\subfigure[$E_\bW$ versus $P$]{\includegraphics[width=0.34\textwidth]{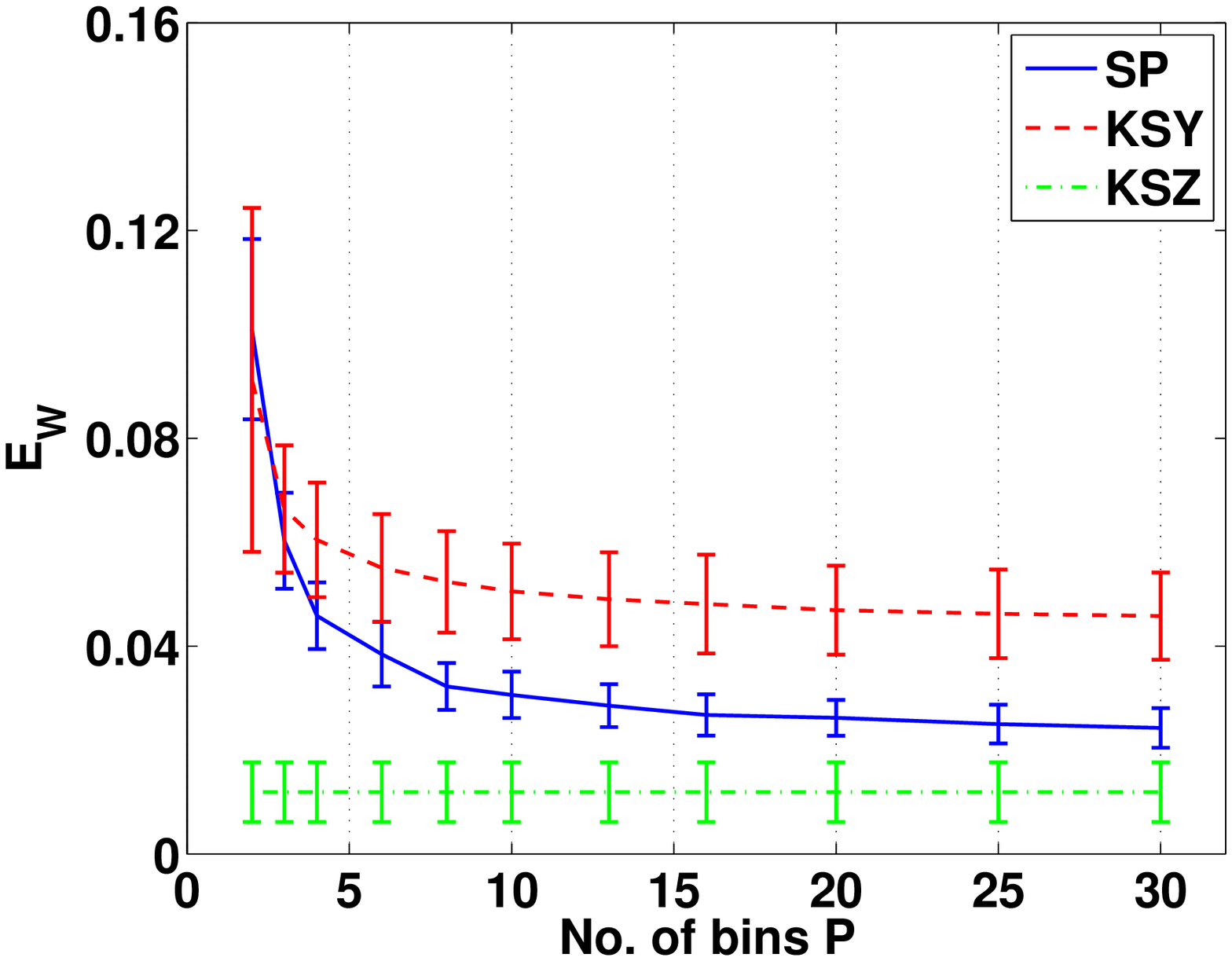}} \hspace{-0.5cm}
\subfigure[$E_\bC$ versus $P$]{\includegraphics[width=0.34\textwidth]{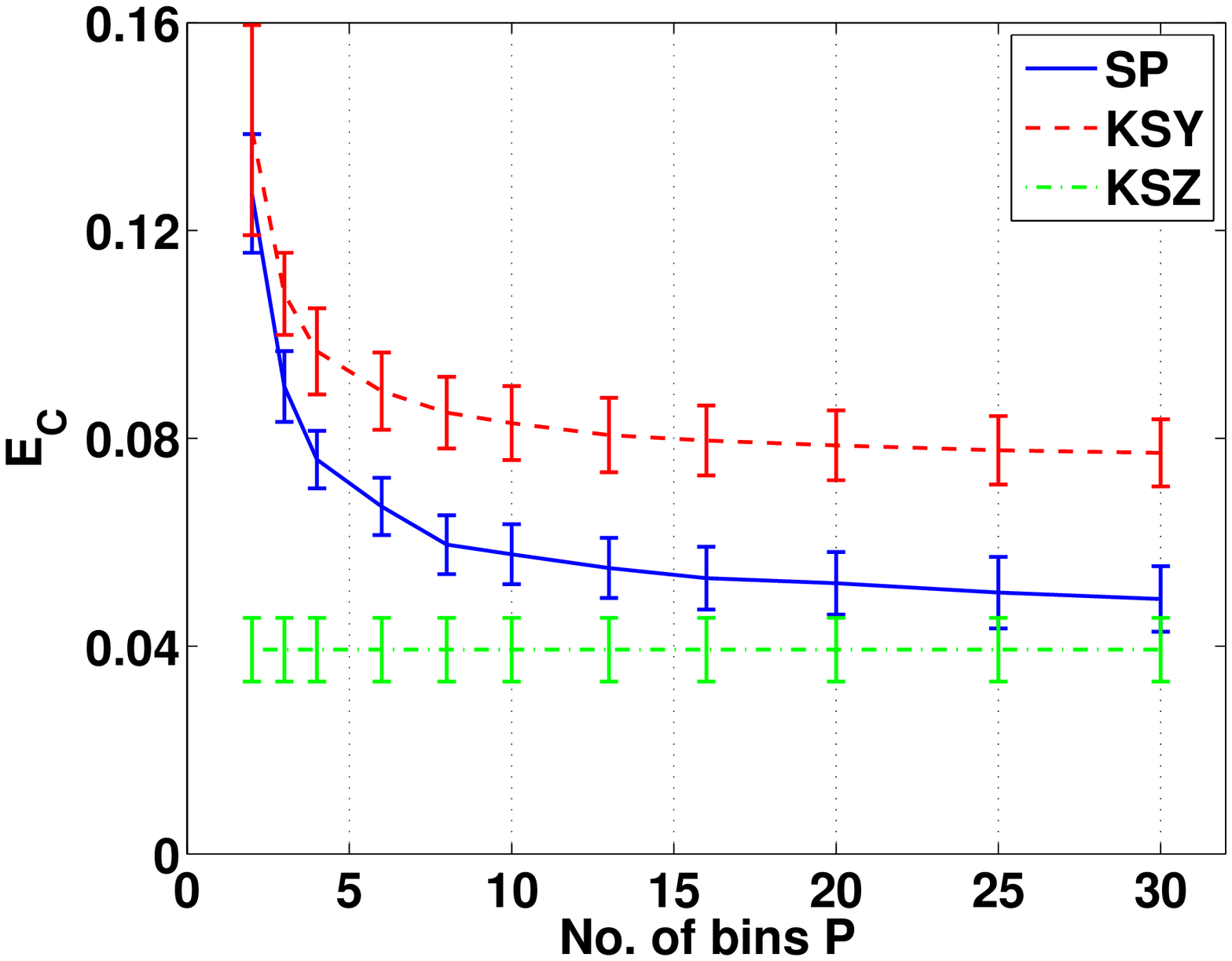}}\hspace{-0.5cm}
\subfigure[$E_{\boldsymbol{\mu}}$ versus $P$]{\includegraphics[width=0.34\textwidth]{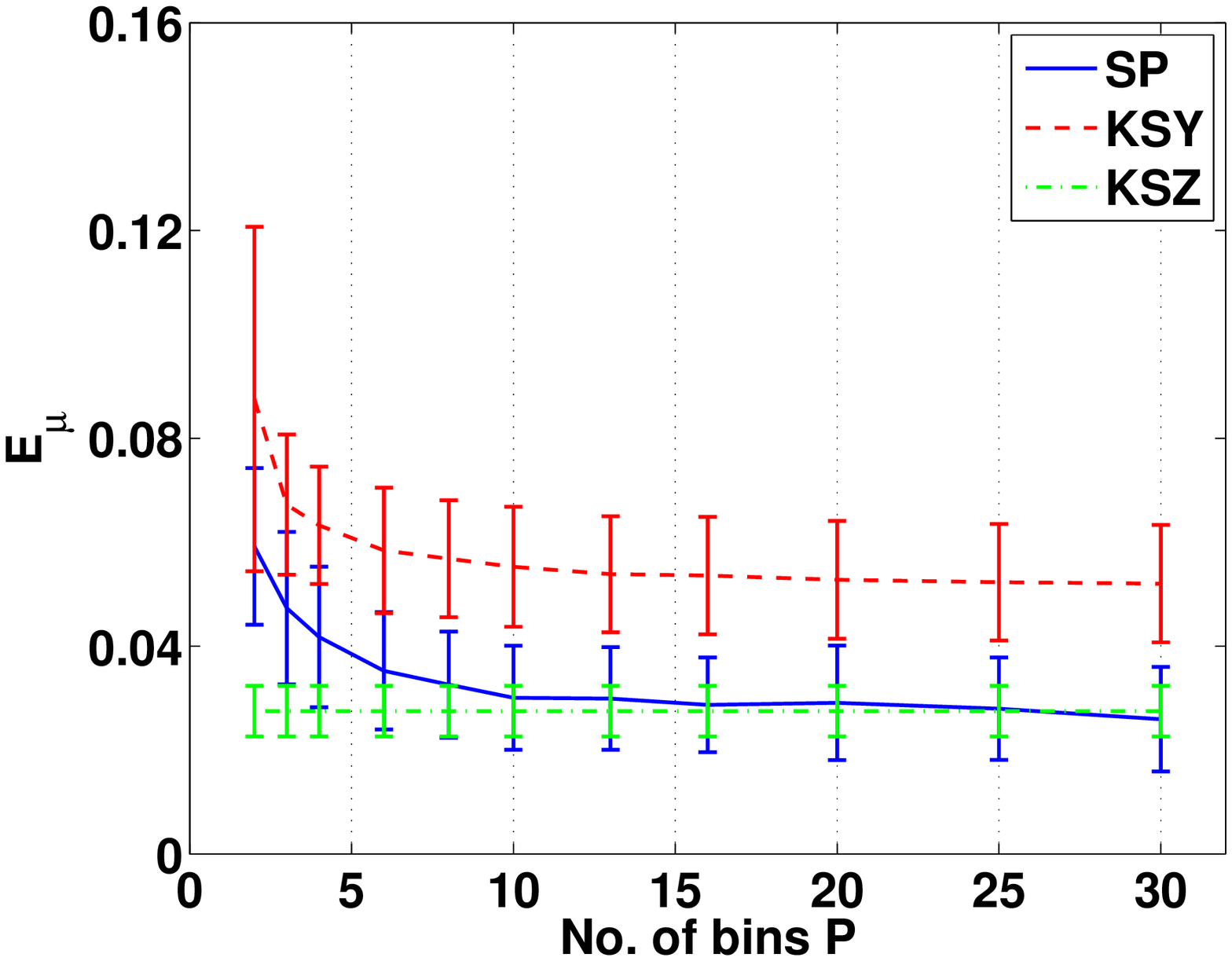}}
\hspace{-0.5cm}
\addtocounter{subfigure}{-3}
\vspace{-0.7cm}
  \caption{Performance comparison of Ordinal SPARFA-M vs. K-SVD$_+$ by varying the number of quantization bins. ``SP'' denotes Ordinal SPARFA-M, ``KSY'' denotes K-SVD$_+$ operating on $\bY$, and ``KSZ'' denotes K-SVD$_+$ operating on $\bZ$ in \fref{eq:qam} (the unquantized data).}
\vspace{0.2cm}
\label{fig:synthlevels}
\end{figure*}

\subsection{Synthetic data}
\label{sec:synth}

\sloppy

In order to show that Ordinal SPARFA-Tag is capable of estimating latent factors based on binary observations, we compare the performance of Ordinal SPARFA-Tag to a non-negative variant of the popular K-SVD dictionary learning algorithm~\cite{ksvd}, referred to as K-SVD$_+$, which we have detailed in~\cite{sparfa}. 
We consider both the case when the precision~$\tau$ is known a-priori and also when it must be estimated.  
In all synthetic experiments, the algorithm parameters $\lambda$ and $\eta$ are selected according to Bayesian information criterion (BIC) \cite{tibsbook}. All experiments are repeated for 25 Monte-Carlo trials.  

\fussy

In all synthetic experiments, we retrieve estimates of all factors, $\widehat{\bW}$, $\widehat{\bC}$, and $\hat{\boldsymbol{\mu}}$.
For Ordinal SPARFA-M and K-SVD$_+$, the estimates $\widehat{\bW}$ and $\widehat{\bC}$ are re-scaled and permuted as in~\cite{sparfa}. We consider the following error metrics:
\begin{align*}
E_\bW \! = \! \frac{\|\bW \! - \! \widehat{\bW}\|_F^2}{\|\bW\|_F^2}, \,\,\, E_\bC \! = \! \frac{\|\bC \! - \! \widehat{\bC}\|_F^2}{\|\bC\|_F^2}, \,\,\,  E_{\boldsymbol{\mu}} \! = \! \frac{\|\boldsymbol{\mu} \! - \! \hat{\boldsymbol{\mu}}\|_2^2}{\|\boldsymbol{\mu}\|_2^2}. 
\end{align*}

\begin{figure*}[tp]
\vspace{-0.3cm}
\begin{floatrow}
\hspace{-.1cm}
\ffigbox{%
\includegraphics[width=0.95\columnwidth]{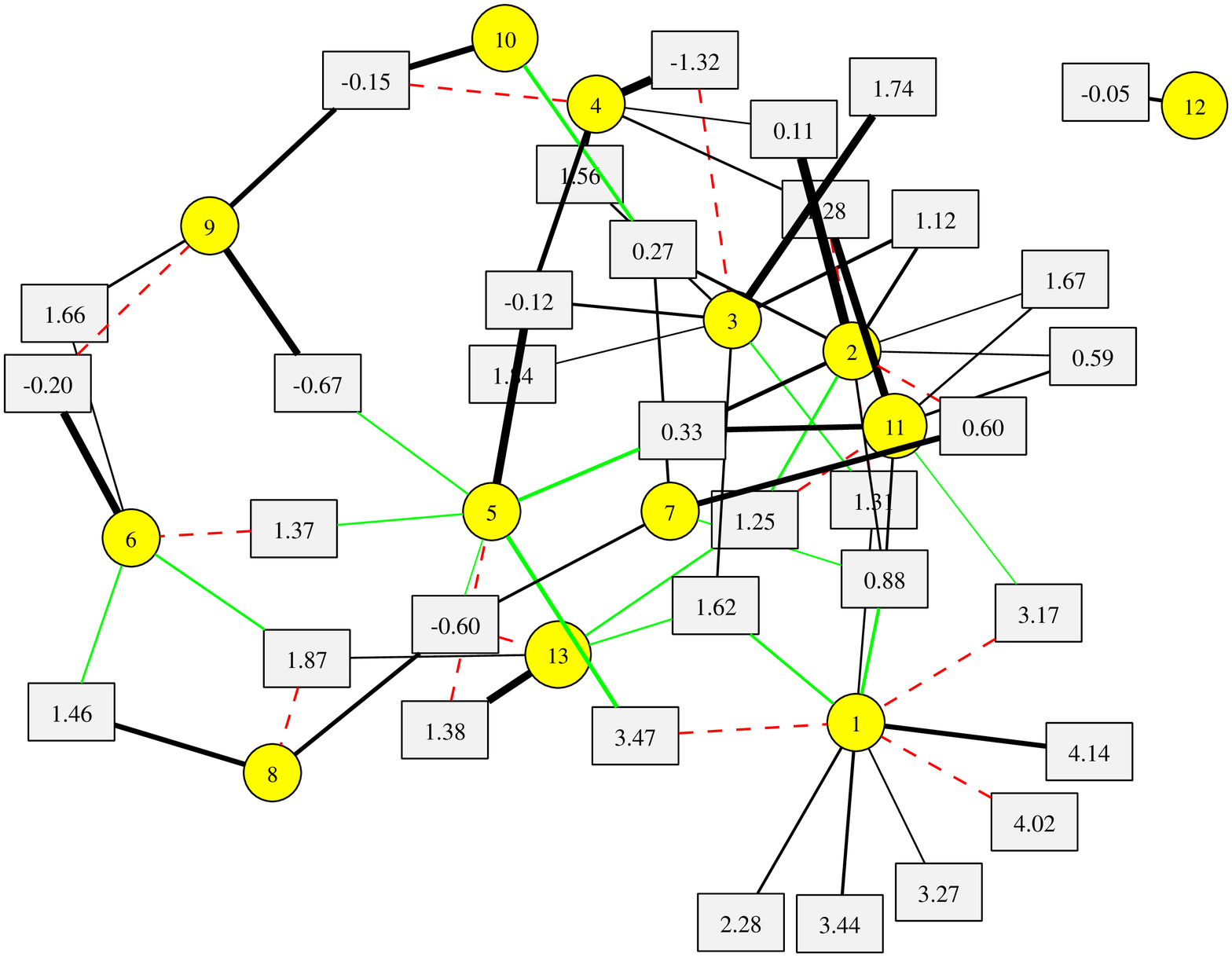} \\[0.1cm]
\scalebox{.7}{%
\begin{tabular}{lllll}
\toprule[0.2em]
Concept 1 & Concept 2 & Concept 3 & Concept 4 & Concept 5\\
\midrule[0.1em]
Arithmetic & Simplifying & Solving & Fractions & Quadratic \\
& expressions & equations & & functions \\
\midrule[0.2em]
 Concept 6 & Concept 7 & Concept 8 & Concept 9 & Concept 10\\ 
\midrule[0.1em]
Geometry & Inequality & Slope & Trigonometry & Limits \\
\midrule[0.2em]
 Concept 11 & Concept 12 & Concept 13\\ 
\midrule[0.1em]
Polynomials & System & Plotting  \\
& equations & functions \\
\bottomrule[0.2em]\\
\end{tabular}}
\vspace{-0.6cm}
}{\caption{Question--concept association graph for a high-school algebra test with $N=99$ users answering $Q=34$ questions. Boxes represent questions; circles represent concepts. We furthermore show the unique tag associated with each concept.}\label{fig:mturkmult}}
\hspace{-0.1cm}
\ffigbox{%
\includegraphics[width=0.85\columnwidth]{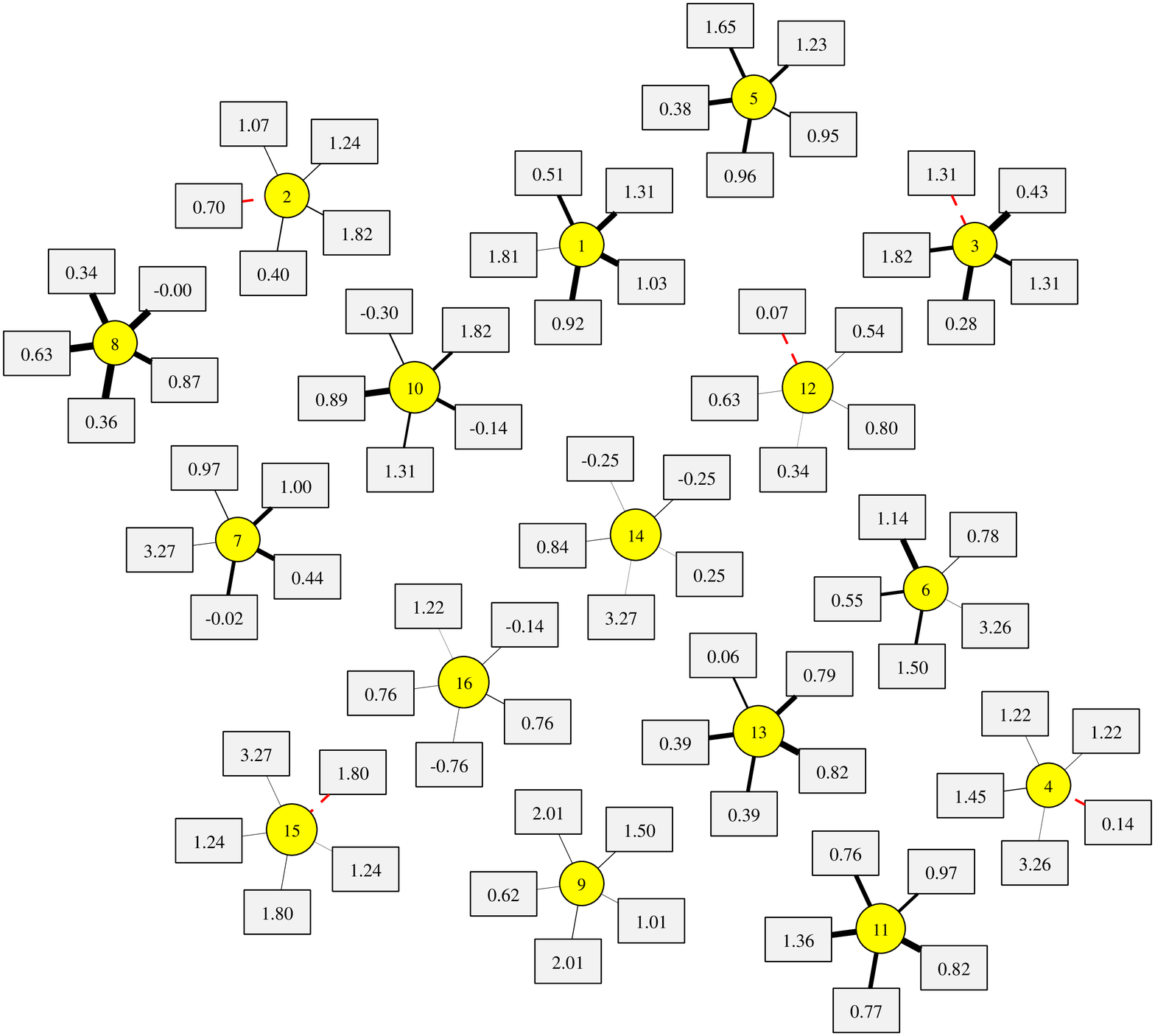} \\[0.0cm]
\scalebox{.65}{%
\begin{tabular}{llllll}
\toprule[0.2em]
Concept 1 & Concept 2 & Concept 3 & Concept 4 & Concept 5 \\
\midrule[0.1em]
Classifying & Properties & Mixtures and & Changes & Uses of  \\
matter & of water & solutions & from heat & energy  \\
\midrule[0.2em]
 Concept 6 & Concept 7 & Concept 8 & Concept 9 & Concept 10 \\ 
\midrule[0.1em]
Circuits and & Forces & Formation of & Changes & Evidence of  \\
electricity & and motion & fossil fuels & to land & the past \\
\midrule[0.2em]
 Concept 11 & Concept 12 & Concept 13 & Concept 14 & Concept 15 & Concept 16\\ 
\midrule[0.1em]
Earth, sun & Alternative & Properties & Earth's & Food & Environmental \\
and moon & energy & of soil & forces & webs & changes \\
\bottomrule[0.2em]\\
\end{tabular}}
\vspace{-0.6cm}
}{\caption{Question--concept association graph for a grade 8 Earth Science course with $N=145$ learners answering $Q=80$ questions ($\bY$ is highly incomplete with only $13.5\%$ entries observed). We furthermore show the unique tag associated with each concept.
}\label{fig:stemsnuc}}
\end{floatrow}
\end{figure*}

We generate the synthetic test data $\bW$, $\bC$, $\boldsymbol{\mu}$ as in \cite[Eq.~10]{sparfa} with $K=5$, $\mu_0=0$, $v_\mu = 1$, $\lambda_k=0.66 \; \forall k$, and \mbox{$\bV_{\!0} = \bI_K$}. $\bY$ is generated according to \fref{eq:qam}, with $P=5$ bins and $\{ \omega_0, \ldots, \omega_5 \} = \{ -\infty, -2.1, -0.64, 0.64, 2.1, \infty \}$, such that the entries of $\bZ$ fall evenly into each bin. 
The number of concepts $K$ for each question is chosen uniformly in $\{1,2,3\}$. 
We first consider the impact of problem size on estimation error in \fref{fig:synthlevels}. To this end, we fix $Q=100$ and sweep $N \in \{50,100,200\}$ for $K = 5$ concepts, and then fix $N=100$ and sweep $Q \in \{50,100,200\}$. 

{\bf Impact of problem size:}
We first study the performance of Ordinal SPARFA-M versus K-SVD$_+$ while varying the problem size parameters $Q$ and $N$.
The corresponding box-and-whisker plots of the estimation error for each algorithm are shown in \fref{fig:synthplots}. 
In \fref{fig:varyn}, we fix the number of questions~$Q$ and plot the errors $E_\bW$, $E_\bC$ and $E_{\boldsymbol{\mu}}$ for the number of learners $N \in \{50,100,200\}$. In \fref{fig:varyq}, we fix the number of learners $N$ and plot the errors $E_\bW$, $E_\bC$ and $E_{\boldsymbol{\mu}}$ for the number of questions $Q \in \{50,100,200\}$.
It is evident that $E_\bW$, $E_\bC$, and $E_{\boldsymbol{\mu}}$ decrease as the problem size increases for all considered algorithms. 
\sloppy
Moreover, Ordinal SPARFA-M has superior performance to K-SVD$_+$ in all cases and for all error metrics.
Ordinal SPARFA-Tag and the oracle support provided versions of K-SVD outperform Ordinal SPARFA-M and K-SVD$_+$. We furthermore see that the variant of Ordinal SPARFA-M without knowledge of the precision~$\tau$  performs as well as knowing~$\tau$; this implies that we can accurately learn the precision parameter directly from data.

\fussy

{\bf Impact of the number of quantization bins:}
We now consider the effect of the number of quantization bins $P$ in the observation matrix $\bY$ on the performance of our algorithms.
We fix $N=Q=100$, $K=5$ and generate synthetic data as before up to $\bZ$ in \fref{eq:qam}. For this experiment, a different number of bins $P$ is used to quantize $\bZ$ into $\bY$. The quantization boundaries are set to $\{\Phi^{-1}(0), \Phi^{-1}(1/P), \ldots, \Phi^{-1}(1) \}$. 
To study the impact of the number of bins needed for Ordinal SPARFA-M to provide accurate factor estimates that are  comparable to algorithms operating with real-valued observations, we also run K-SVD$_+$ directly on the $\bZ$ values (recall $\fref{eq:qam}$) as a base-line. 
\sloppy
Figure~\ref{fig:synthlevels} shows that the performance of \mbox{Ordinal SPARFA-M} consistently outperforms K-SVD$_+$. 
We furthermore see that all error measures decrease by about half when using $6$ bins, compared to $2$ bins (corresponding to binary data). Hence, ordinal \mbox{SPARFA-M} clearly outperforms the conventional SPARFA model \cite{sparfa}, when ordinal response data is available.
As expected, Ordinal SPARFA-M approaches the performance of K-SVD$_+$ operating directly on~$\bZ$ (unquantized data) as the number of quantization bins~$P$ increases. 

\fussy

\subsection{Real-world data}
\label{sec:real}

We now demonstrate the superiority of Ordinal SPARFA-Tag compared to regular SPARFA as in  \cite{sparfa}. In particular, we show the advantages of using tag information directly within the estimation algorithm and of imposing a nuclear norm constraint on the matrix $\bC$.
For all experiments, we apply Ordinal SPARFA-Tag to the graded learner response matrix $\bY$ with oracle support information obtained from instructor-provided question tags. The parameters $\lambda$ and $\eta$ are selected via cross-validation.

{\bf Algebra test:}
\label{sec:realmturk}
We analyze a dataset from a high school algebra test carried out on Amazon Mechanical Turk \cite{mechturkwebsite}, a crowd-sourcing marketplace. 
The dataset consists of $N=99$ users answering $Q=34$ multiple-choice questions covering topics such as geometry, equation solving, and visualizing function graphs. The questions were manually labeled with a set of 13 tags.  The dataset is fully populated, with no missing entries. A domain expert manually mapped each possible answer to one of $P = 4$ bins, i.e., assigned partial credit to each choice as follows: totally wrong ($p=1$), wrong ($p=2$), mostly correct ($p=3$), and correct ($p=4$). 

\sloppy

Figure~\ref{fig:mturkmult} shows the question--concept association map estimated by Ordinal SPARFA-Tag using the Frobenius norm constraint $\|\bC\|_F \leq \eta$.
Circles represent concepts, and squares represent questions (labelled by their intrinsic difficulties $\mu_i$). 
Large positive values of $\mu_i$ indicate easy questions; negative values indicate hard questions. 
Connecting lines indicate whether a concept is present in a question; thicker lines represent stronger question--concept associations. 
Black lines represent the question--concept associations estimated by Ordinal SPARFA-Tag, corresponding to the entries in $\bW$ as specified by $\Gamma$. Red, dashed lines represent the ``mislabeled'' associations (entries of $\bW$ in $\Gamma$) that are estimated to be zero. Green solid lines represent new discovered associations, i.e., entries in $\bW$ that were not in~$\Gamma$ that were discovered by Ordinal SPARFA-Tag.

\fussy

By comparing \fref{fig:mturkmult} with \cite[Fig.~9]{sparfa}, we can see that Ordinal SPARFA-Tag provides unique concept labels, i.e., one tag is associated with one concept; this enables precise interpretable feedback to individual learners, as the values in $\bC$ represent directly the tag knowledge profile for each learner. 
This tag knowledge profile can be used by a PLS to provide targeted feedback to learners.
The estimated question--concept association matrix can also serve as useful tool to domain experts or course instructors, as they indicate missing and inexistent tag--question associations. 

{\bf Grade 8 Earth Science course:}
\label{sec:stems}
As a second example of Ordinal SPARFA-Tag, we analyze a Grade~$8$ Earth Science course dataset \cite{stemwebsite}. 
This dataset contains $N=145$ learners answering $Q=80$ questions and is highly incomplete (only $13.5\%$ entries of $\bY$ are observed). The matrix $\bY$ is binary-valued; domain experts labeled all questions with $16$ tags. 

\sloppy

The result of Ordinal SPARFA-Tag with the nuclear norm constraint $\|\bC\|_* \leq \eta$ on $\bY$ is shown in  \fref{fig:stemsnuc}. The estimated question--concept associations mostly matches those pre-defined by domain experts. Note that our algorithm identified some question--concept associations to be non-existent (indicated with red dashed lines). Moreover, no new associations have been discovered, verifying the accuracy of the pre-specified question tags from domain experts. Comparing to the question--concept association graph of the high school algebra test in \fref{fig:mturkmult}, we  see that for this dataset, the pre-specified tags represent disjoint knowledge components, which is indeed the case in the underlying question set. 
Interestingly, the estimated concept  matrix $\bC$ has rank~$3$; note that we are estimating $K=16$ concepts. This observation suggests that all learners can be accurately represented by a linear combination of only 3 different ``eigen-learner'' vectors. Further investigation of this clustering phenomenon  is part of on-going research.

\fussy

\subsection{Predicting unobserved learner responses}
\label{sec:pred}

\sloppy

\begin{figure}[t]
\includegraphics[width=.85\textwidth]{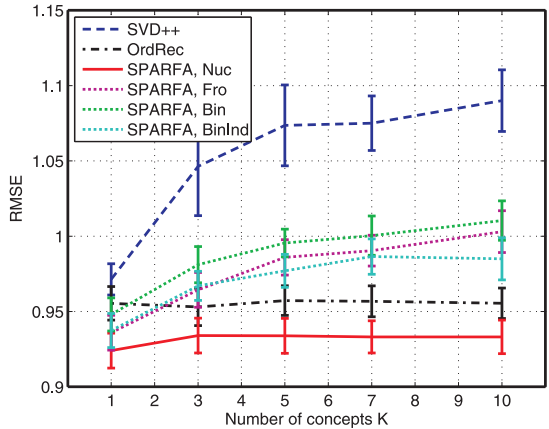}
\vspace{-0.3cm}
  \caption{Prediction performance on the Mechanical Turk algebra test dataset. We compare the collaborative filtering methods SVD$++$ and OrdRec to various Ordinal SPARFA-M based methods: ``Nuc'' uses the nuclear norm constraint, ``Fro'' uses the Frobenius norm constraint, ``Bin'' and ``BinInd'' learn the bin boundaries, whereas ``Bin'' learns one set of bin boundaries for the entire dataset and ``BinInd'' learns individual bin boundaries for each question.}
\vspace{-0.1cm}
\label{fig:pred}
\end{figure}

We now compare the prediction performance of ordinal \mbox{SPARFA-M} on unobserved learner responses against state-of-the-art collaborative filtering techniques: (i) SVD$++$ in~\cite{svdplus}, which treats ordinal values as real numbers, and (ii) \mbox{OrdRec} in \cite{ordpred}, which relies on an ordinal logit model.
We compare different variants of Ordinal SPARFA-M: (i) optimizing the precision parameter, (ii) optimizing a set of bins for all learners, (iii) optimizing a set of bins for each question, and (iv) using the nuclear norm constraint on $\bC$.
We consider the Mechanical Turk algebra test, hold out 20\% of the observed learner responses as test sets, and train all algorithms on the rest. The regularization parameters of all algorithms are selected using $4$-fold cross-validation on the training set. Figure~\ref{fig:pred} shows the root mean square error (RMSE) $\sqrt{\frac{1}{\mid \bar{\Omega}_\text{obs} \mid} \textstyle \sum_{i,j:(i,j)\in \bar{\Omega}_\text{obs}} \| \hat{Y}_{i,j} - Y_{i,j} \|_2^2 }$ where $\hat Y_{i,j}$ is the predicted score for $Y_{i,j}$, averaged over 50 trials. 

Figure~\ref{fig:pred} demonstrates that the nuclear norm variant of Ordinal SPARFA-M outperforms OrdRec, while the performance of other variants of ordinal SPARFA are comparable to \mbox{OrdRec}. \mbox{SVD$++$} performs worse than all compared methods, suggesting that the use of a probabilistic model considering ordinal observations enables accurate predictions on unobserved responses. 
We furthermore observe that the variants of Ordinal SPARFA-M that optimize the precision parameter or bin boundaries deliver almost identical performance. 

We finally emphasize that Ordinal SPARFA-M not only delivers superior prediction performance over the two state-of-the-art collaborative filtering techniques in predicting learner responses, but it also provides interpretable factors, which is key in educational applications.

\fussy

\section{Related Work} \label{sec:rlwork}

A range of different ML algorithms have been applied in educational contexts.  
Bayesian belief networks have been successfully used to probabilistically model and analyze learner response data in order to trace learner concept knowledge and estimate question difficulty (see, e.g., \cite{kt,gregk1,estdiff,woolf08}). Such models, however, rely on predefined question--concept dependencies (that are not necessarily accurate), in contrast to the framework presented here that estimates the dependencies solely from data. 

\sloppy
Item response theory (IRT) uses a statistical model to analyze and score graded question response data \cite{lordirt,mirt}.
Our proposed statistical model shares some similarity to the Rasch model \cite{rasch}, the additive factor model \cite{afm}, learning factor analysis \cite{stamperbp,lfa}, and the instructional factors model~\cite{ifm}. 
These models, however, rely on pre-defined question features, do not support disciplined algorithms to estimate the model parameters solely from learner response data, or do not produce interpretable estimated factors.
Several publications have studied factor analysis approaches on learner responses \cite{qmatrix,qprobit,viettwo}, but treat learner responses as real and deterministic values rather than ordinal values determined by statistical quantities. 
Several other results have considered probabilistic models in order to characterize learner responses \cite{predfact,logitfa}, but consider only  binary-valued responses and cannot be generalized naturally to ordinal data.

\fussy

While some ordinal factor analysis methods, e.g., \cite{ordpred}, have been successful in predicting missing entries in datasets from ordinal observations, our model enables interpretability of the estimated factors, due to (i) the additional structure imposed on the learner--concept matrix (non-negativity combined with sparsity) and (ii) the fact that we associate unique tags to each concept within the estimation algorithm.

\section{Conclusions}
\label{sec:conclusions}

\sloppy

We have significantly extended the SPARse Factor Analysis (SPARFA) framework of \cite{sparfa} to exploit (i) ordinal learner question responses and (ii) instructor generated tags on questions as oracle support information on the question--concept associations. 
We have developed a computationally efficient new algorithm to compute an approximate solution to the associated ordinal factor-analysis problem.
Our proposed \emph{Ordinal SPARFA-Tag} framework not only estimates the strengths of the pre-defined question--concept associations provided by the instructor but can also discover new associations. 
Moreover, the algorithm is capable of imposing a nuclear norm constraint on the learner concept knowledge matrix, which achieves better prediction performance on unobserved learner responses than state-of-the-art collaborative filtering techniques, while improving the interpretability of the estimated concepts relative to the user-defined tags.

The Ordinal SPARFA-Tag framework enables a PLS to provide readily interpretable feedback to learners about their latent concept knowledge. 
The tag-knowledge profile can, for example, be used to make personalized recommendations to learners, such as recommending remedial or enrichment material to learners according to their tag (or concept) knowledge status.  Instructors also benefit from the capability to discover new question--concept associations underlying their learning materials. 

\section{Acknowledgments}

Thanks to Daniel Calderon for administering the Algebra test on Amanzon's Mechanical Turk.
This work was supported by the National Science Foundation under Cyberlearning grant IIS-1124535, the Air Force Office of Scientific Research under grant FA9550-09-1-0432, the Google Faculty Research Award program, and the Swiss National Science Foundation under grant PA00P2-134155.


\bibliographystyle{abbrv}
\bibliography{sparfaclustbib}

\end{document}